\documentclass[10pt,final,compsoc]{IEEEtran}
\usepackage{times}
\usepackage{epsfig}
\usepackage{graphicx}
\usepackage{amsmath}
\usepackage{amssymb}
\usepackage{makecell}
\usepackage{slashbox}
\usepackage{colortbl}
\usepackage{array}
\usepackage{multirow}
\usepackage[table]{xcolor}
\usepackage[none]{hyphenat}
\usepackage{color}
\usepackage{ragged2e}

\ifCLASSOPTIONcompsoc
  \usepackage[nocompress]{cite}
\else
  \usepackage{cite}
\fi

\ifCLASSINFOpdf
\else
\fi
\begin{document}
%
\title{A Framework for Human Pose Estimation in Videos}

\author{Dong~Zhang~\IEEEmembership{}
        and Mubarak~Shah~\IEEEmembership{}
\IEEEcompsocitemizethanks{\IEEEcompsocthanksitem D. Zhang and M. Shah are with the Department
of Computer Science, University of Central Florida, Orlando,
FL, 32817.\protect\\
E-mail: \{dzhang,shah\}@eecs.ucf.edu}
\thanks{}}

%
%

\markboth{}%
{Shell \MakeLowercase{\textit{et al.}}: Bare Demo of IEEEtran.cls for Computer Society Journals}
%



\IEEEtitleabstractindextext{%
\begin{abstract}
\justify
In this paper, we present a method to estimate a sequence of human poses in unconstrained videos. We aim to demonstrate that by using temporal information, the human pose estimation results can be improved over image based pose estimation methods. In contrast to the commonly employed graph optimization formulation, which is NP-hard and needs approximate solutions, we formulate this problem into a unified two stage tree-based optimization problem for which an efficient and exact solution exists. Although the proposed method finds an exact solution, it does not sacrifice the ability to model the spatial and temporal constraints between body parts in the frames; in fact it models the {\em symmetric}  parts better than the existing methods. The proposed method is based on two main ideas: `Abstraction' and `Association' to enforce the intra- and inter-frame body part constraints without inducing extra computational complexity to the polynomial time solution. Using the idea of `Abstraction', a new concept of `abstract body part' is introduced to conceptually combine the symmetric body parts and model them in the tree based body part structure. Using the idea of `Association', the optimal tracklets are generated for each abstract body part, in order to enforce the spatiotemporal constraints between body parts in adjacent frames. A sequence of the best poses is inferred  from the abstract body part tracklets through the tree-based optimization. Finally, the poses are refined by limb alignment and refinement schemes. We evaluated the proposed method on three publicly available video based human pose estimation datasets, and obtained dramatically improved performance compared to the state-of-the-art methods.
\end{abstract}

\begin{IEEEkeywords}
Human Pose Estimation, Motion Detection, Object Detection
\end{IEEEkeywords}}

\maketitle

\IEEEdisplaynontitleabstractindextext

%
\IEEEpeerreviewmaketitle

\IEEEraisesectionheading{\section{Introduction}\label{sec_introduction}}

%
%
%
%

\IEEEPARstart{H}{uman} pose estimation is crucial for many computer vision applications, including human computer interaction, activity recognition and video surveillance. It is a very challenging problem due to the large appearance variance, non-rigidity of the human body, different viewpoints, cluttered background, self occlusion etc. Recently, a significant progress has been made in solving the human pose estimation problem in unconstrained single images \cite{Yang2011,Ramakrishna2014,Toshev2014,Zhang2015}; however, human pose estimation in videos \cite{Park2011,Ramakrishna2013,Cherian2014} is a relatively new and promising problem and needs significant improvement. Obviously, single image based pose estimation method can be applied to each video frame to get an initial pose estimation, and a further refinement through frames can be applied to make the pose estimation consistent and more accurate. However, due to the innate complexity of video data, the problem formulations of most video based human pose estimation methods are very complex (usually NP-hard), therefore, approximate solutions have been proposed to solve them which result in sub-optimal solutions. Furthermore, most of the existing methods  model body parts as a tree structure and these methods tend to suffer from double counting issues \cite{Ramakrishna2014} (which means symmetric parts, for instance left and right ankles, are easily to be  mixed together). This paper aims to formulate the video based human pose estimation problem in a different manner, which makes the problem solvable in polynomial time with an exact solution, and also effectively enforces the spatiotemporal constraints between body parts (which will handle the double counting issues).

\begin{figure}
\begin{center}
\includegraphics[width=3.2in]{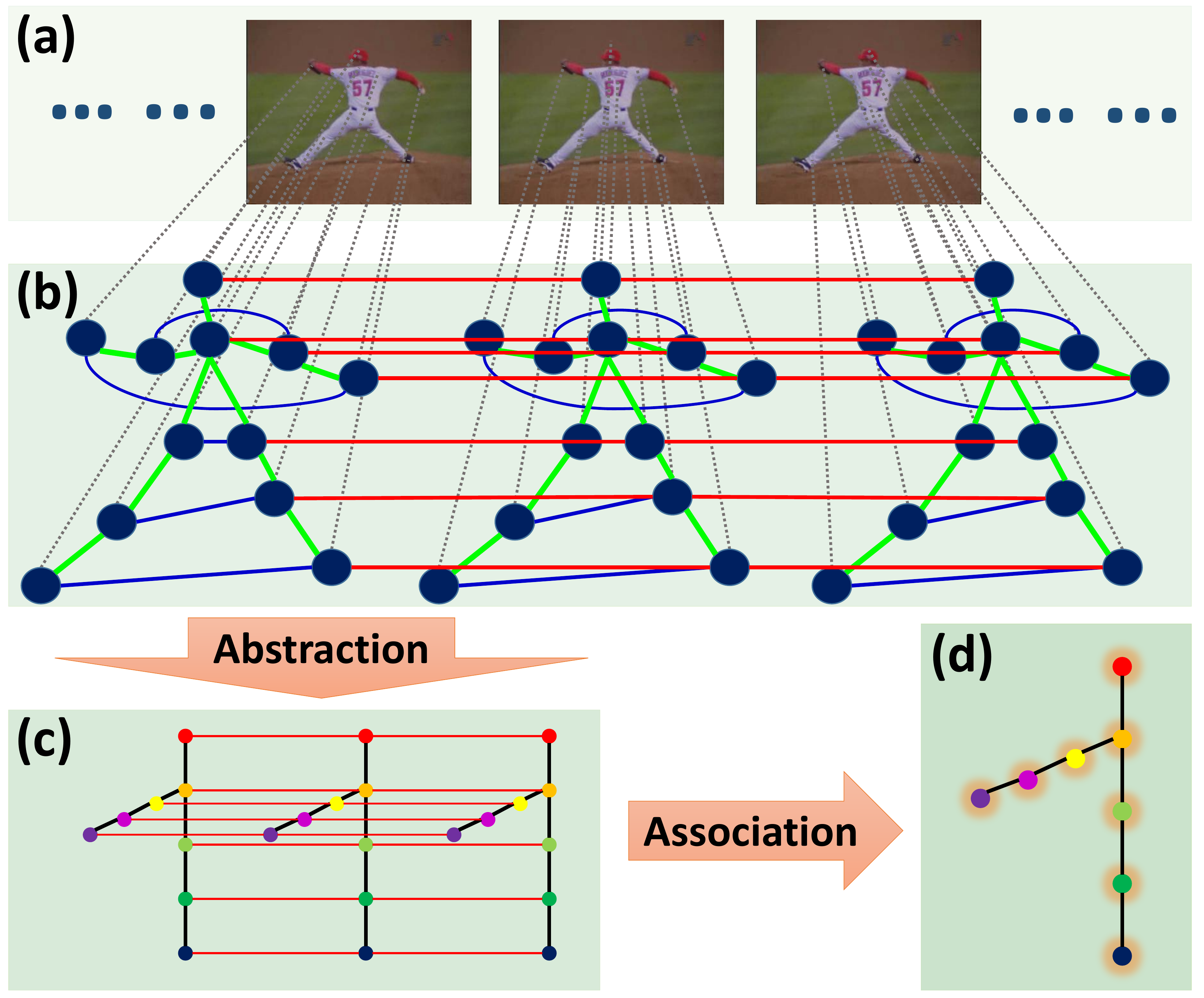}
\end{center}
   \caption{An abstract high-level illustration of the proposed method aiming at removing simple cycles from the commonly employed graph optimization framework for video based human pose estimation problem. All of the above graphs are relational graphs for the problems. In (a), a few sample frames of a video are shown. In (b), each body part in each frame is represented by a node. Green and blue edges represent relationships between different body parts in the same frame (green ones are commonly used edges in the literature, and blue ones are important edges for symmetric parts); red edges represent the consistency constraints for the same body part in adjacent frames. Note that this is only an illustration and not all edges are shown.
   In the `Abstraction' stage, symmetric parts are combined together, and the simple cycles within each single frame are removed (shown in (c)); and in the `Association' stage, the simple cycles between adjacent frames are removed (shown in (d)). }
   \label{fig_loopy_graph}
\end{figure}

\begin{figure}
\begin{center}
\includegraphics[width=3.2in]{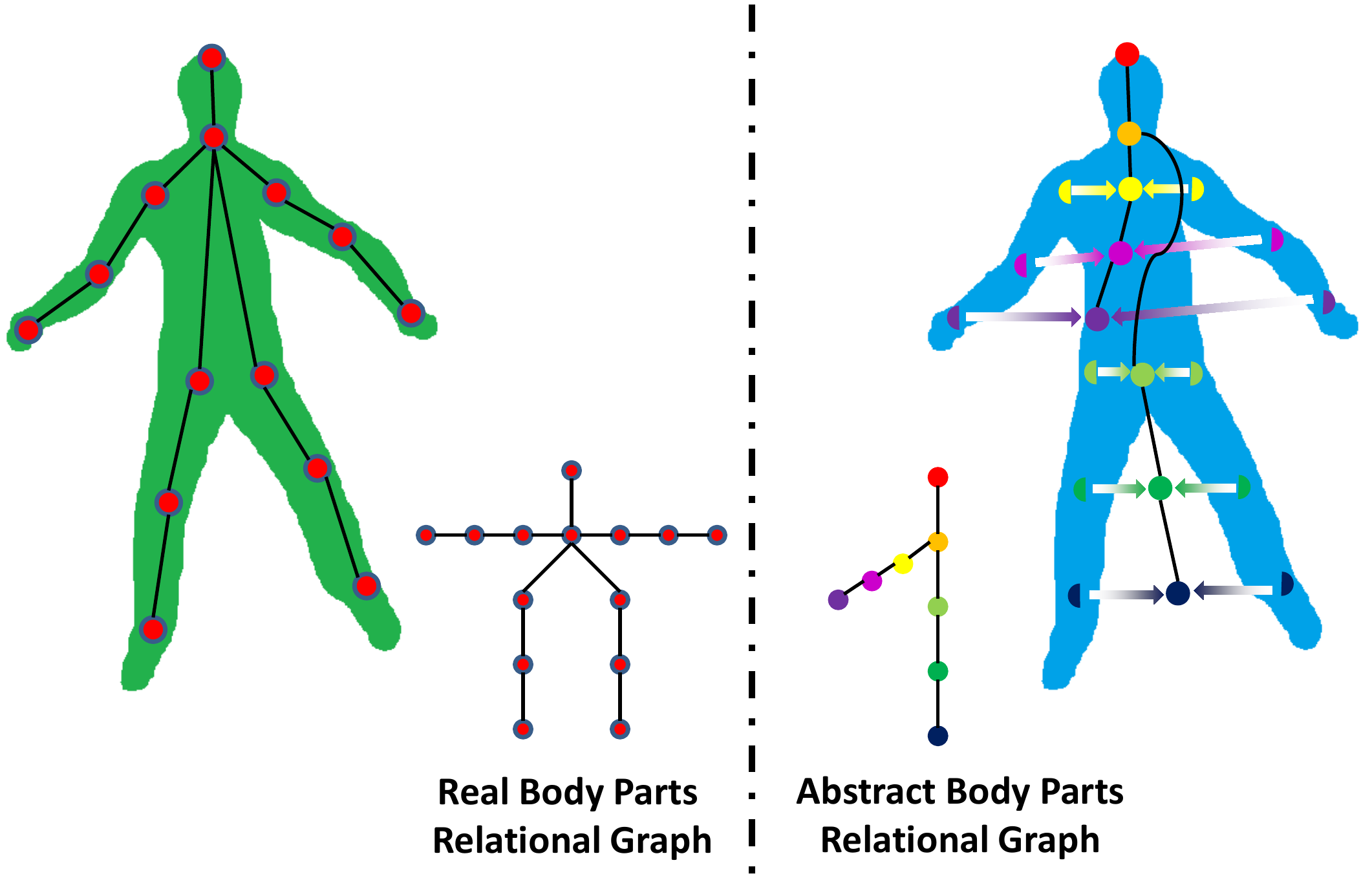}
\end{center}
   \caption{Real body parts vs. abstract body parts. The left side shows a commonly used body part definitions in the literature, and we call these body parts (nodes) `real body parts', and the graph `real body part relational graph'. The right side shows the proposed new definition of body parts, basically we combine a pair of symmetric body parts to be one body part, we call these body parts (nodes) `abstract body parts', since the parts are some abstract concepts of parts, not real body parts, and the graph as `abstract body part graph'.}
   \label{fig_definitions_of_body_parts}
\end{figure}

One commonly employed methodology for human pose estimation in videos is the graph optimization formulation. There are two types of such formulation. \textbf{One type} of this formulation \cite{Park2011} is to generate several human pose hypotheses in each frame and select one best hypothesis from each frame, while making sure they are consistent throughout the video. The inference in this approach is very efficient, however, due to the large variations of pose configurations, it is very difficult to get good poses with all body parts correctly estimated. Therefore, \textbf{another type} of such formulation \cite{Ramakrishna2013,Tokola2013,Cherian2014} was introduced to handle each body part separately (please see Fig.\ref{fig_loopy_graph}(b)). In this formulation, hypotheses are generated for each body part in every frame, and following the spatial constraints between body parts in each frame and temporal consistency of appearances and locations between adjacent frames, the goal is to optimally select the best part hypotheses for each part from all the frames together. This formulation is desirable, since it is able to expand enough diverse human pose configurations, and also, it is able to model spatiotemporal constraints between body parts effectively. Despite all the benefits of this formulation, it is an NP-hard problem due to the underlying loopy graph structure (i.e. there are too many simple cycles in the graph; e.g. the simple cycles in Fig.\ref{fig_loopy_graph}(b) induced by the green, blue and red edges) \cite{Ramakrishna2013}. Several methods were proposed to attack this NP-hard problem in different ways. To reduce the complexity induced by inter-frame simple cycles, Tokola {\em et.al} \cite{Tokola2013} proposed a tracking-by-selection framework which tracks each body part separately and combines them at a latter stage. Authors in \cite{Cherian2014} proposed an approximate method which focuses on less-certain parts in order to reduce the complexity. Ramakrishna {\em et. al} \cite{Ramakrishna2013} introduced a method which takes symmetric parts into account and proposed an approximate solution to handle the loopy graph. And in \cite{Sapp2011}  the original model is decomposed into many sub-models which are solvable since the sub-models have a tree-based structure. All of the above methods are insightful, however, none of them has simultaneously exploited the important constraints between body parts (e.g. symmetry of parts) and has an efficient exact solution.

Based on the discussion above, the major issue is: How to exploit the spatial constraints between the body parts in each frame and temporal consistency through frames to the greatest possible extent, with an efficient exact solution? Since the inference of a tree-based optimization problem has a polynomial time solution \cite{Yang2011,Zhang2013}, the issue becomes (please refer to Fig.\ref{fig_loopy_graph}): How to formulate the problem in order to model the useful spatial and temporal constraints between body parts in the frames without inducing simple cycles? We propose two key ideas to tackle this issue, which approximate the original fully connected model into a simplified tree-based model. The first idea is {\bf Abstraction}: in contrast to the standard tree representation of body parts, we introduce a new concept, {\em abstract body parts}, to conceptually combine the symmetric body parts (please refer to Fig.\ref{fig_definitions_of_body_parts}, and details are introduced in Section \ref{sec_body_parts}). It takes the advantage of the symmetric nature of the human body parts without inducing simple cycles into the formulation. The second idea is {\bf Association}, using which we generate optimal tracklets for each abstract body part to ensure the temporal consistency. Since each abstract body part is processed separately, it does not induce any temporal simple cycles into the graph.

The proposed method is different from the state-of-the-art methods \cite{Ramakrishna2013,Tokola2013,Cherian2014,Sapp2011} in the following ways: \cite{Ramakrishna2013} exploits the symmetric nature of body parts, however, the problem is formulated as a multi-target tracking problem with mutual exclusions, which is NP-complete and only approximate solutions can be obtained by relaxation; the method in \cite{Tokola2013} is designed to remove the temporal simple cycles from the graph shown in Fig.\ref{fig_loopy_graph}(b) to track upper body parts, however, the employed junction tree algorithm will have much higher computational complexity if applied to full-body pose estimation, since there are many more simple cycles induced by symmetric body parts; compared to \cite{Cherian2014}, the proposed method has no temporal simple cycles; and in contrast to \cite{Sapp2011}, it can model symmetric body part structure more accurately rather than settling for the sub-models. Therefore, the proposed method ensures both spatial and temporal constraints, without inducing any simple cycles into the formulation and an exact solution can be efficiently found by dynamic programming.

The organization of the rest of the paper is as follows. After discussing related work in Section \ref{sec_related}, we introduce the proposed method in Section \ref{sec_method}. We present experimental results in Section \ref{sec_experiments} and conclude the paper in Section \ref{sec_conclusion}.

\begin{figure*}[htb]
\begin{center}
\includegraphics[width=7in]{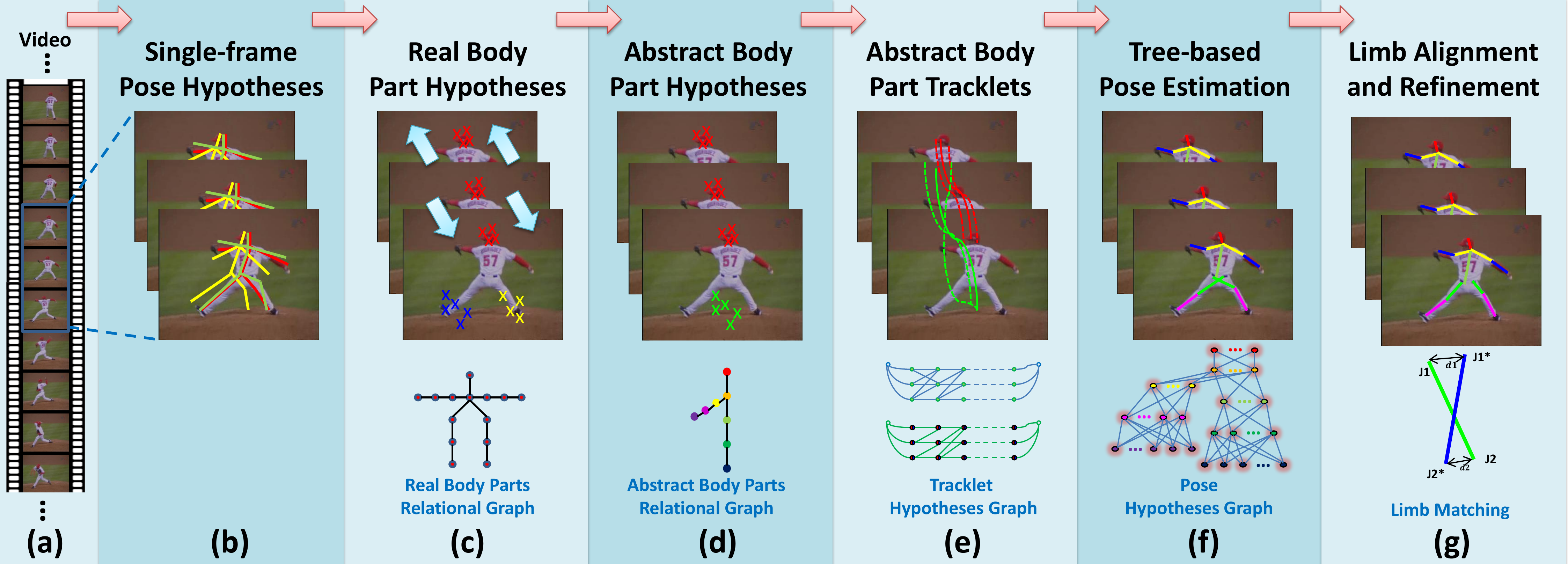}
\end{center}
   \caption{An outline of the proposed method. (a) shows the original video frames; in (b), pose hypotheses in each frame are generated by N-Best method \cite{Park2011} or DCNNGM \cite{Chen2014}; in (c), by using the results from (b), real body part hypotheses are generated for each body part in each frame and propagated to the adjacent frames; in (d), real body parts are combined into abstract body parts and the hypotheses are also combined accordingly in order to remove the intra-frame simple cycles (i.e. the simple cycles with blue and green edges in Fig.\ref{fig_loopy_graph}(b)); in (e), tracklets are generated for abstract body parts (including single body parts and coupled body parts) using the abstract body part hypotheses generated in (d); in (f), the pose hypotheses graph is built, each node is a tracklet corresponding to the abstract body part, and the best pose estimation is obtained by selecting the best hypotheses for the parts from the graph; and in (g), the correct limbs are inferred by limb alignment and refinement schemes.}
   \label{fig_framework}
\end{figure*}

\section{Related Work}
\label{sec_related}

A large body of work in human pose estimation have been reported in recent years. Early works are focused on human pose estimation and tracking in controlled environment \cite{Sigal2010}; there is also some important work using depth images \cite{Shotton2013}. Single image based human pose estimation \cite{Yang2011,Toshev2014,Dantone2013,Wang2013} in unconstrained scenes has  progressed dramatically in the last a few years; however, video based human pose estimation in unconstrained scenes is still in a very early stage, and some pioneer research \cite{Ramakrishna2013,Park2011,Tokola2013,Cherian2014} has been conducted only recently.

For image based human pose estimation in unconstrained scenes, most work has been focused on pictorial structure models \cite{Andriluka2009,Andriluka2010,Johnson2010,Ramanan2006} for quite long time and the performance has been promising. In \cite{Yang2011}, a flexible mixture-of-parts model was proposed to infer the pose configurations, which showed very impressive results. A new scheme was introduced in \cite{Johnson2011} to handle a large number of training samples which resulted in significant increase in pose estimation accuracy. Authors in \cite{Simo-Serra2012,Yu2013,Ramakrishna2012} attempted to estimate 3D human poses from a single image. The high order dependencies of body parts are exploited in \cite{Lan2005,Jiang2011,Jiang2008,Sun2012,Sigal2006,Wang2008,Karlinsky2012,Tian2012,Pishchulin2013,Ramakrishna2014}.  The authors in \cite{Tian2012} proposed a hierarchical spatial model with an exact solution, while \cite{Pishchulin2013} achieves this by defining a conditional model, and \cite{Ramakrishna2014} employs an inference machine to explore the rich spatial interactions among body parts. A novel, non-linear joint regressor model was proposed in \cite{Dantone2013}, which handles typical ambiguities of tree based models quite well. More recently, deep learning\cite{Toshev2014,Tompson2014,Ouyang2014,Chen2014} has also been introduced for human pose estimation.

For video based human pose estimation in unconstrained scenes, some early research adopted the tracking-by-detection framework \cite{Andriluka2008,Mori2004,Ramanan2005}.  More recently, some methods \cite{Cherian2014,Tokola2013,Zuffi2013,Fragkiadaki2013,Rohrbach2012,Sapp2011} have mainly focused on upper body pose estimation and other methods \cite{Park2011,Ramakrishna2013} have focused on full body pose estimation. In \cite{Park2011},  many pose candidates are generated in each frame, and the most consistent ones which have high detection scores are selected through the frames.  A tracking-by-selection framework is proposed in \cite{Tokola2013} to simplify the graph optimization problem, which makes the exact inference possible. In \cite{Cherian2014},   the poses are decomposed into limbs, and recomposed together to obtain pose estimations in the video. The authors in \cite{Sapp2011} decompose the full model of body parts into many tree-based sub-models which enables them to get the exact inference for the sub-models. Ramakrishna {\em et al.}\cite{Ramakrishna2013} model the symmetric structures of the body parts and proposed an effective approximate solution to the problem.


\section{Tree-based Optimization for Human Pose Estimation in Videos}
\label{sec_method}

We formulate the video based human pose estimation problem into a {\em unified} tree-based optimization framework, which can be solved efficiently by dynamic programming. With a reference to the major steps in Fig.\ref{fig_framework}, we introduce the general notions of relational graph and hypothesis graph, and related problem formulation and solutions in Section \ref{sec_graphs}; we discuss the new concept: `abstract body parts' in comparison to `real body parts' in section \ref{sec_body_parts} and show how to generate body part hypotheses in each frame in Section \ref{sec_single_frame_hypotheses}; we introduce tracklets generation in Section \ref{sec_single_part_tracklets} and \ref{sec_coupled_part_tracklets}; we show how to extract the optimal poses in Section \ref{sec_optimal_pose_estimation}, and we introduce the limb alignment and refinement schemes in Section \ref{sec_limb_alignment_and_refinement}.

The main steps of the method (as show in Fig.\ref{fig_framework} (b - g)) are: \textbf{1)} for each frame, generate many pose hypotheses by the N-Best method \cite{Park2011} or DCNNGM \cite{Chen2014}; \textbf{2)} based on step 1, generate hypotheses for each real body part and prorogate them to adjacent frames (We use the term \emph{real body parts} to represent body parts which are commonly used in the literature, and \emph{abstract body
parts} as a new concept which will be introduced later in this paper to facilitate the formulation of the proposed method); \textbf{3)} combine the symmetric real body part hypotheses and obtain the abstract body part hypotheses; \textbf{4)} build the tracklet hypotheses graph for each abstract body part and select the top trackles for each; \textbf{5)} build the pose hypotheses graph for the abstract body part tracklets and select the best pose configuration. \textbf{6)} use limb alignment and refinement schemes to infer the correct limb configurations.

\subsection{Relational Graph vs. Hypothesis Graph}
\label{sec_graphs}

\begin{figure}
\begin{center}
\includegraphics[width=3.2in]{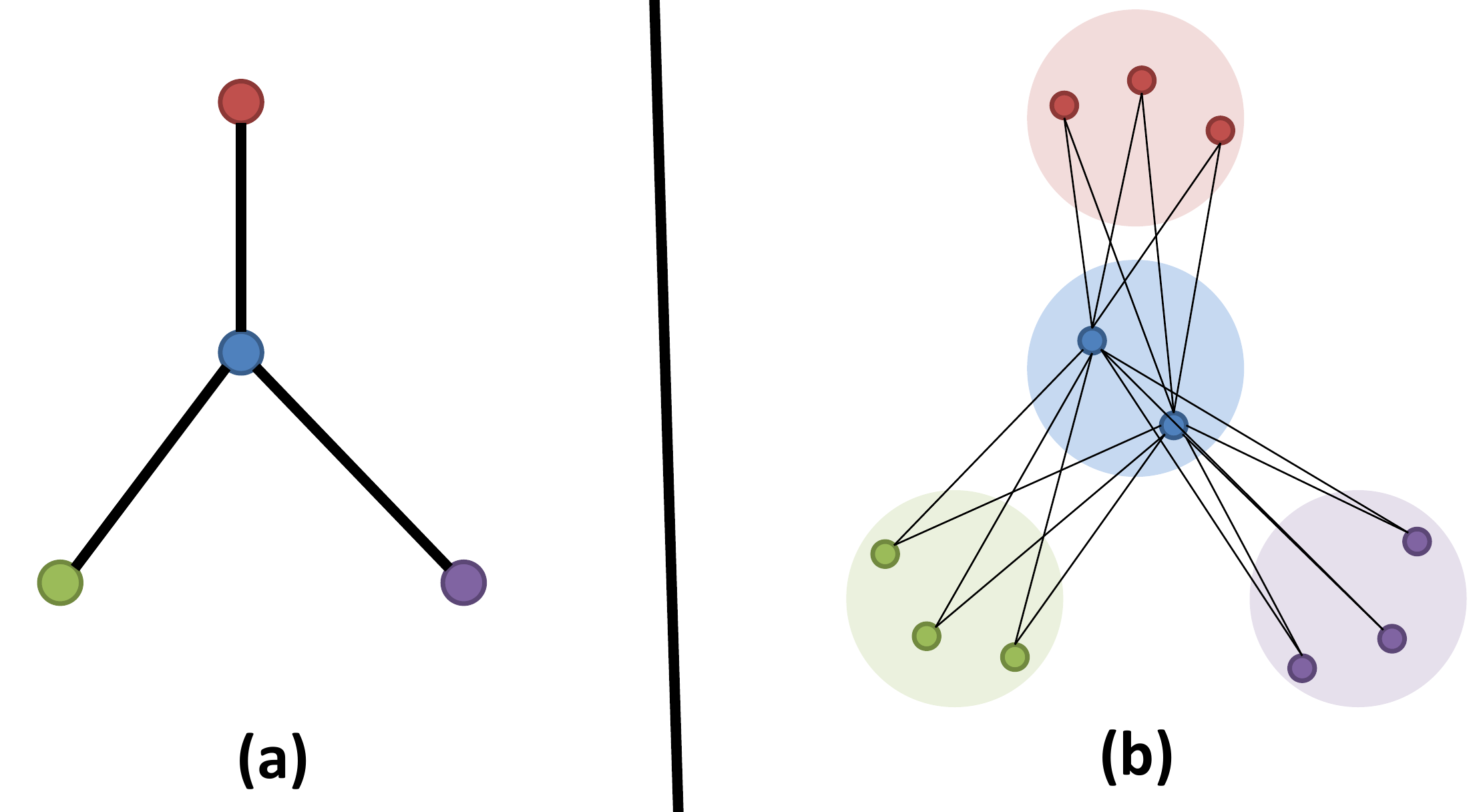}
\end{center}
   \caption{Relational Graph vs. Hypothesis Graph. (a) shows the relational graph for a problem. (b) shows the corresponding hypothesis graph. In this example, each entity has three hypotheses and corresponding to the structure of (a), edges are added between the hypotheses which have relationship.}
   \label{fig_graphs}
\end{figure}

In computer vision, and several other disciplines, many problems can be abstracted as follows. Assume there is a set of entities $\mathcal{E} = \{e^i|_{i=1}^N\}$, where each of them can stay in only one of the many states $\mathcal{S} = \{s^k|_{k=1}^M\}$, with the unary scoring functions $\{\Phi(e^i,s^k)|e^i \in \mathcal{E}, s^k \in \mathcal{S} \}$, which gives the likelihood that an entity $e^i$ stays in state $s^k$. And there is a binary compatibility function for each pair of entities $\{\Psi(e^i,e^j,s^k,s^l) | e^i, e^j \in \mathcal{E}, s^k, s^l \in \mathcal{S}\}$, which represents the compatibility of entity $e^i$ in state $s^k$ and entity $e^j$ in state $s^l$. The goal is to decide the best states for each entity such that all of them have high unary scores and are also compatible with each other. This problem can be modeled as a graph optimization problem formulated by relational and hypothesis graphs, which is described next.

A relational graph, $G_r = (V_r,E_r)$, represents the relationship of a set of entities which are represented by entity nodes $\{v_r^{i}\left.\right|_{i=1}^{|V_r|}\}$, and the relationships between pairs of entities are represented by edges $E_r$. Fig.\ref{fig_graphs}(a) shows a simple relational graph which has 4 entities and 3 edges to represent their relations (more examples of relational graph are shown in Fig.\ref{fig_definitions_of_body_parts}, Fig.\ref{fig_tracklet_hypothesis_graph}(a) and Fig.\ref{fig_pose_hypothesis_graph}(a)). Corresponding to a relational graph $G_r$, a hypothesis graph, $G_h = (V_h,E_h)$, can be built. Fig.\ref{fig_graphs}(b) is the hypothesis graph for the relational graph in Fig.\ref{fig_graphs}(a). For an entity node $v_r^i$ in $V_r$, a \emph{group} of hypothesis nodes $V_{h(i)} = \{ v_{h(i)}^k \left.\right|_{k=1}^{|V_{h(i)}|}\}$ are generated to form the hypothesis graph, so $V_h = \underset{i=1}{\overset{|V_r|}{\bigcup}} V_{h(i)}$. The hypothesis nodes represent the possible states of each entity, and in this paper they represent possible locations of body parts. Hypothesis edges, $E_h = \{(v_{h(i)}^k,v_{h(j)}^l)|v_{h(i)}^k \in V_{h(i)},v_{h(j)}^l \in V_{h(j)}, (v_r^i,v_r^j) \in E_r\}$, are built between each pair of hypothesis nodes from different \emph{group}s following the structure of $G_r$. An unary weight, $\Phi$, can be assigned to each hypothesis node, which measures the likelihood for the corresponding entity to be in the state of this hypothesis node; and a binary weight, $\Psi$, can be assigned to each hypothesis edge, which measures the compatibility of the pair of hypothesis nodes connected by the edge. Examples of hypothesis graph are shown in Fig.\ref{fig_tracklet_hypothesis_graph}(b,c) and Fig.\ref{fig_pose_hypothesis_graph}(b). The methodology is to select one hypothesis node for each entity, in order to maximize the combined unary and binary weights. This is a graph optimization problem and the general form is NP-hard; however, if the relational graph is a tree (including the degenerated case of a single branch), the problem is no longer NP-hard and efficient dynamic programming based polynomial time solutions exist.

For a tree-based relational graph, $G_r$, and the corresponding hypothesis graph, $G_h$, the objective function for a set of arbitrary selected nodes
(following the structure of $G_r$)
$s = \{s^i|_{i=1}^{|V_r|},s^i \in V_h\}$ is:
\begin{equation}
\label{eqn_objective_function_tree}
    \mathcal{M}(s) = \sum_{s^i \in V_h}{\Phi(s^i)} + \lambda \cdot \sum_{(s^i,s^j) \in E_h}{\Psi(s^i,s^j)},
\end{equation}
in which $\lambda$ is the parameter for adjusting the binary and unary weights, and the goal is to maximize $\mathcal{M}(s)$: $s^* = \arg \max_{s}(\mathcal{M}(s))$. Let the algorithm proceed from the leaf nodes to the root, and let $\mathcal{F}(i,k)$ be the maximum achievable combined unary and binary weights of $k$th hypothesis for $i$th entity. $\mathcal{F(\cdot,\cdot)}$ satisfies the following recursive function:
\begin{align}
    \mathcal{F}(i,k) &= \Phi(v_{h(i)}^k) + \nonumber \\
    &\sum_{v_r^j \in kids(v_r^i)}{\max_{l}{\left(\lambda \cdot \Psi(v_{h(i)}^k,v_{h(j)}^l) + \mathcal{F}(j,l)\right)}}.
\end{align}
Based on this recursive function, the problem can be solved efficiently by dynamic programming, with a computation complexity of $\mathcal{O}(|V_r| \cdot N)$, in which $N$ is the max number of hypotheses for each node in $V_r$.

\subsection{Real Body Parts vs. Abstract Body Parts}
\label{sec_body_parts}

We use the term {\em real body parts} to represent body parts which are commonly used in the literature. And we use {\bf abstract body parts}, which is a new concept introduced in this paper to facilitate the formulation of the proposed method (as shown in Fig.\ref{fig_definitions_of_body_parts}). In contrast to the  real body part definitions, there are two types of the abstract body parts in this paper: \textbf{single part} and \textbf{coupled part}. \textbf{Single parts} include \emph{HeadTop} and \emph{HeadBottom}. \textbf{Coupled parts} include \emph{Shoulder}, \emph{Elbow}, \emph{Hand}, \emph{Hip}, \emph{Knee} and \emph{Ankle}. Note that, for coupled parts, we use one part to represent two symmetric real body parts, for instance \emph{Ankle} is employed to represent the abstract part which is actually the combination of the \emph{left} and \emph{right} ankles. The design of abstract body parts is to remove simple cycles for the body part relational graph, but keep the ability to model the symmetric body parts. For example, in Fig.\ref{fig_loopy_graph}(b), in each frame, the green and blue edges are designed to model the body part relationships, but there are many simple cycles in a single frame. After introducing the abstract body parts in Fig.\ref{fig_loopy_graph}(c), the symmetric parts are combined, thus there is no simple cycle anymore in each frame. There are still simple cycles between frames, which will be handled by the abstract body part tracklets in Section \ref{sec_single_part_tracklets} and \ref{sec_coupled_part_tracklets}.

\subsection{Body Part Hypotheses in a Single Frame}
\label{sec_single_frame_hypotheses}

There are several ways to generate body part hypotheses in each video frame. In \cite{Zhang2015}, N-Best method \cite{Park2011} is used to generate the hypotheses. In this paper, although N-Best method \cite{Park2011} is still an essential step, we also explore generation of body part hypotheses by deep learning method (to be specific, we are using deep convolutional neural networks with graphical model (DCNNGM) \cite{Chen2014}), in our experiments. For the  sake of completeness, we  briefly explain hypotheses generation using N-Best method \cite{Park2011} and DCNNGM method \cite{Chen2014} in this section.

\subsubsection{N-Best Hypotheses}
\label{sec_NBest_hypotheses}

N-Best human pose estimation approach \cite{Park2011} is applied to each video frame to generate $N$ best full body pose hypotheses. $N$ is usually a large number (normally $N>300$). And for each real body part, the body part hypotheses are body part locations extracted from the N-best poses. The body part hypotheses are sampled by an iterative non-maximum suppression (NMS) scheme based on the detection score map. Detection score is a combination of max-marginal \cite{Park2011} and foreground score,
\begin{equation}
\label{eqn_detection_score}
    \Phi_{s}(p) = \alpha \Phi_{M}(p) + (1-\alpha) \Phi_{F}(p),
\end{equation}
where $\Phi_{s}$ is the detection score, $\Phi_{M}$ is the max-marginal derived from \cite{Park2011}, $\Phi_{F}$ is the foreground score obtained by the background subtraction \cite{Elgammal2000}, and $p$ is the location of the body part.

\subsubsection{DCNNGM Hypotheses}
\label{sec_DCNNGM_hypotheses}

\begin{figure}
\begin{center}
\includegraphics[width=3.5in]{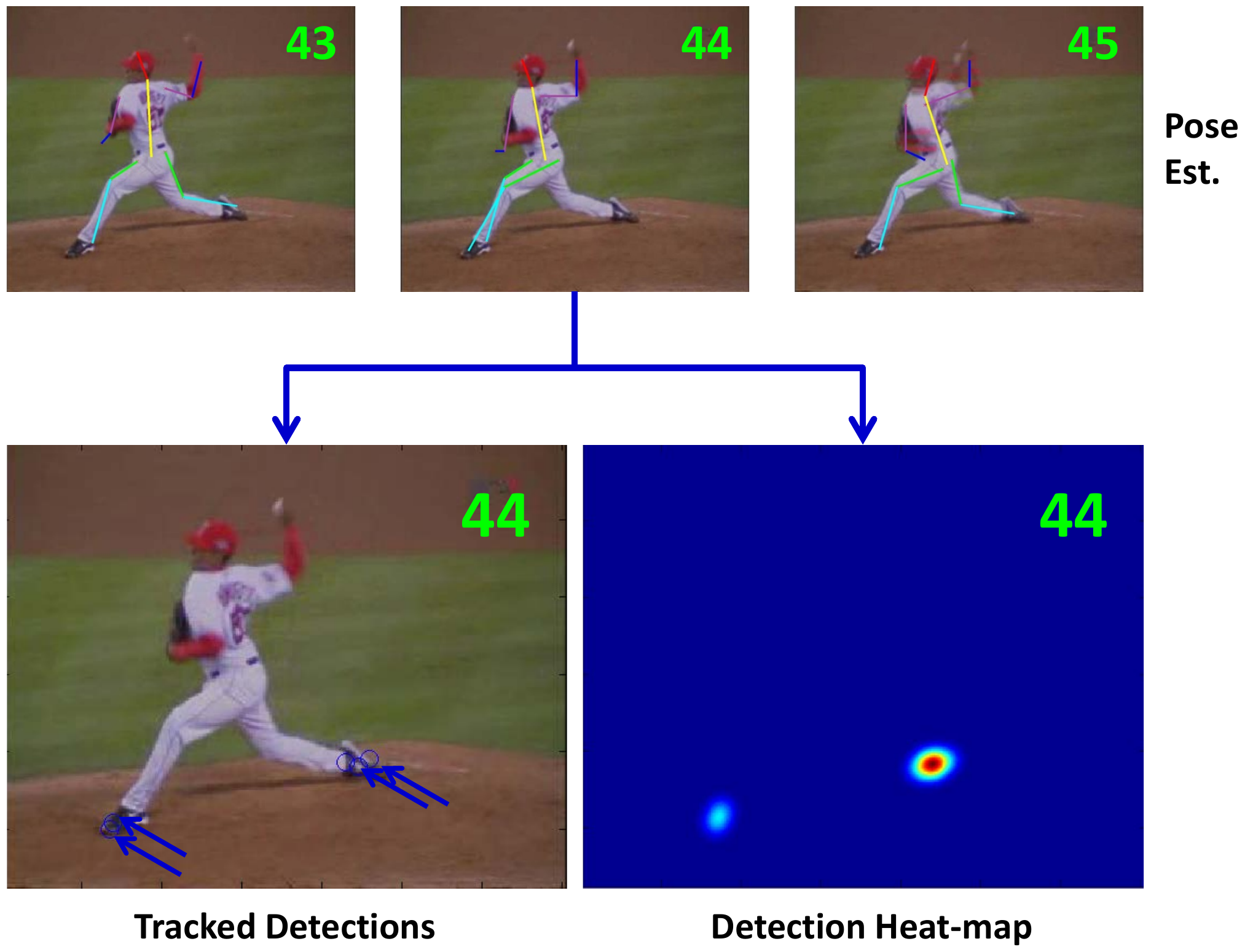}
   \caption{An Illustration for DCNNGM Pose Hypotheses Generation. The top row shows the pose estimation results using DCNNGM \cite{Chen2014}, and the bottom row shows the tracked ``right ankle" for the neighboring frames and the combined detection heat-map. In this example, in frame 44, the right leg was not correctly localized; however, in some of its neighboring frames, the right leg was localized correctly. Therefore, by tracking using neighboring frames, we can generate good detection hypotheses for 'right ankle' and thus a better detection map.}
   \label{fig_pose_hypotheses_generation}
\end{center}
\end{figure}

Compared to N-Best method \cite{Park2011}, deep learning methods for human pose estimation \cite{Toshev2014,Ouyang2014,Chen2014,Tompson2014} have shown better performance in single image pose estimation, and we want to take advantage of this. In this paper, we employ the deep convolutional neural networks with graphical model approach (DCNNGM) \cite{Chen2014} to generate pose hypotheses. We chose this approach because its code is available and it is one of the state-of-the-art deep learning methods for pose estimation, however, we believe other state-of-the-art methods \cite{Toshev2014,Ouyang2014,Tompson2014} can also get comparable, or even better results. There are two problems to solve before we can employ DCNNGM in our framework: \textbf{1)} Although the method can process a small image in a reasonable amount of time, it is still prohibitive to handle videos which have many frames and the frame sizes are relatively large (e.g. $1280 \times 720$). \textbf{2)} The method does not output multiple hypotheses for each frame.

We solve the first problem by a pre-processing step: N-Best method \cite{Park2011} is employed to estimate the poses in each video frame, and the top estimations are used to infer the rough bounding boxes of the people in each frame. The bounding boxes are cropped and input to DCNNGM to estimate the poses and results are mapped back to the original video frames. N-Best method \cite{Park2011} is an essential step here, but not the only option, and any other pose estimation method (e.g. \cite{Yang2011}) or human detection method (e.g. \cite{Felzenszwalb2008}) may be employed here with proper adjustments.

We solve the second problem by tracking the pose estimation from neighboring frames. This process not only generates more pose hypotheses in each video frame, but also makes the estimation more robust and consistent (please refer to Fig. \ref{fig_pose_hypotheses_generation} for an illustration). In every frame, each body part is tracked backward and forward to neighboring frames by a simple NCC (normalized cross correlation) tracker. NCC tracker is employed here due to its simplicity and efficiency, and since the parts are only tracked for a few frames. We decided not to employ heavy-weight trackers not to overkill the problem, although it is possible that the results may be improved slightly by employing some fancier tracker (e.g. \cite{Kalal2012,Wang2015}).

In each frame and for every body part, by tracking poses from neighboring frames, the outcome will be several candidate locations (one from the detection in the current frame, and others are from the tracking results from other frames). An 2D Gaussian is applied to each location and an detection heat-map is obtained (please refer to Fig.\ref{fig_pose_hypotheses_generation}). By sampling many locations from the detection heat-map, several body part hypotheses are generated. The values of the detection heat-map are used as $\Phi_{M}(p)$ in Eqn.\ref{eqn_detection_score}.

\subsubsection{Abstract Body Part Hypotheses}

After generating the body part hypotheses, either using the method in Section \ref{sec_NBest_hypotheses} or Section \ref{sec_DCNNGM_hypotheses}, the abstract body part hypotheses are obtained: the abstract body part hypotheses for a single part are the same as its corresponding real body part hypotheses. And the abstract body part hypotheses for a coupled part are the permutation of its corresponding left and right body part hypotheses.

\begin{figure}
\begin{center}
\includegraphics[width=3.2in]{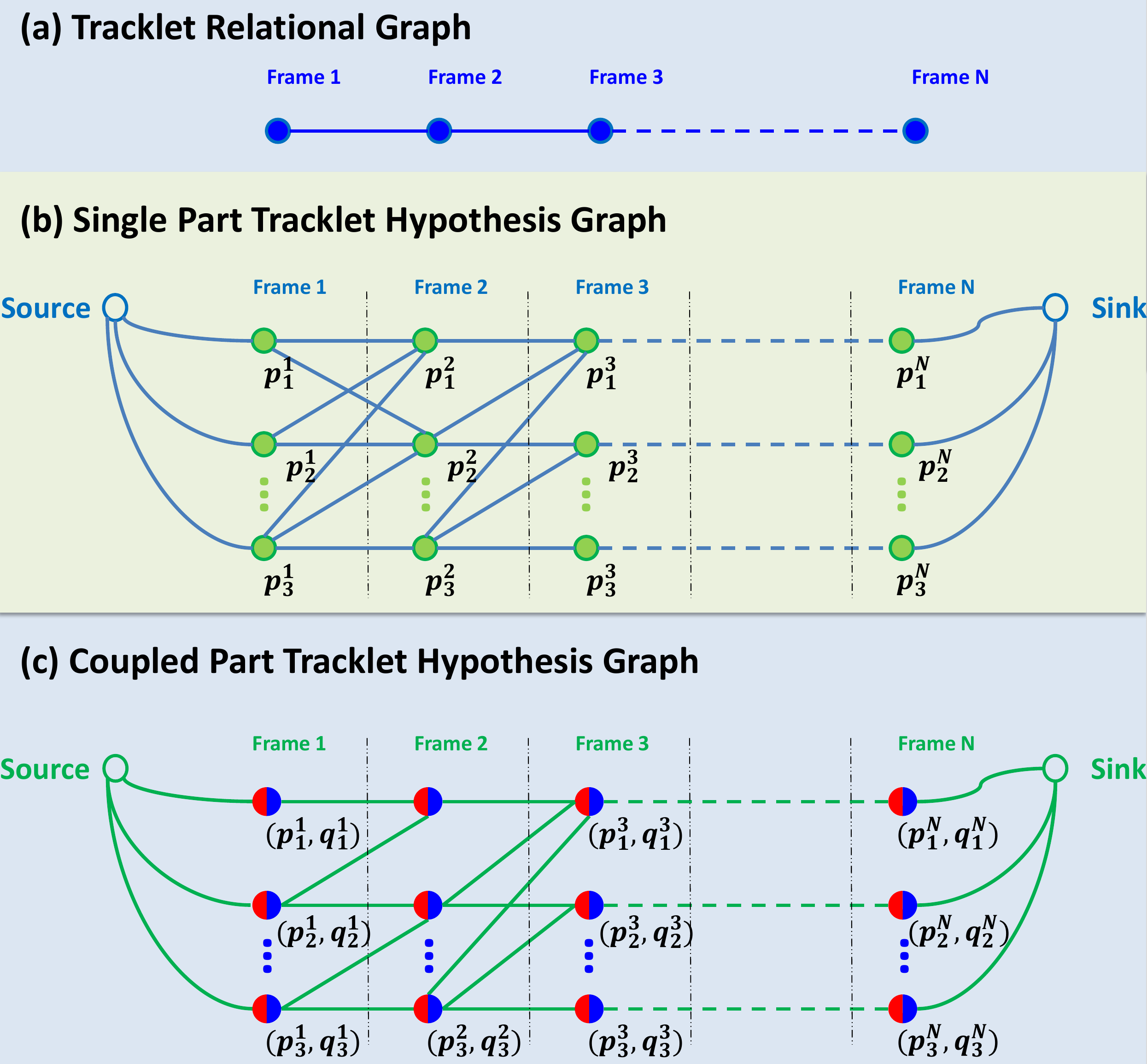}
\end{center}
   \caption{\textbf{Tracklet graphs}. (a) Shows the relational graph for the abstract body part tracklet generation. (b) Shows the tracklet hypothesis graph for single body parts. Each node represents one hypothesis location of the body part in a specific frame, and edges show the similarity between the connected body part hypotheses in adjacent frames. (c) Shows the tracklet hypothesis graph for coupled parts. Each node represents a coupled body part hypothesis, which is the combination of the corresponding symmetric body parts (that is why each node is colored into two halves). The edges represent the similarities between connected coupled body parts in adjacent frames. Note that, (b) and (c) are only illustrations, and for simplicity, not all edges are shown. }
   \label{fig_tracklet_hypothesis_graph}
\end{figure}

\subsection{Single Part Tracklets}
\label{sec_single_part_tracklets}

Based on the abstract body part hypotheses generated in Section \ref{sec_single_frame_hypotheses}, we want to obtain several best single part and coupled part tracklets through the video frames. The problem is how to select one hypothesis from each frame ensuring that they have high detection scores and are consistent throughout the frames. Following the definitions in Section \ref{sec_graphs}, the relational graph for this problem is shown in Fig.\ref{fig_tracklet_hypothesis_graph}(a), and the hypothesis graphs for single parts and coupled parts are shown in Fig.\ref{fig_tracklet_hypothesis_graph} (b) and (c) respectively.

With the single part hypotheses, a single part tracklet hypothesis graph is built (Fig.\ref{fig_tracklet_hypothesis_graph}(b)) for each single part (\emph{headTop} and \emph{headBottom}). Each node represents a single part hypothesis and the detection score $\Phi_{s}(p)$ in Eqn.\ref{eqn_detection_score} is used to assign  the node  an unary weight. Edges are added between every pair of nodes from the adjacent frames. Binary weights are assigned to the edges which represent similarities between hypotheses in adjacent frames. The binary weight is defined as a combination of optical flow predicted spatial distance and Chi-square distance of HOG features as follows:

\begin{equation}
\label{eqn_single_part_binary_weight}
    \Psi_{s}(p^f,p^{f+1}) = \exp(-\frac{\chi^2(\Upsilon(p^{f}),\Upsilon(p^{f+1}) \cdot \| \hat{p}^f - p^{f+1} \|_2^2 )}{\sigma^2}),
\end{equation}
where $p^f$ and $p^{f+1}$ are two arbitrary hypotheses from frames $f$ and $f+1$, $\Upsilon(p)$ is the HOG feature vector centered at location $p$, $\hat{p}^f$ is the optical flow predicted location for $p^f$ in frame $f+1$, and $\sigma$ is a parameter. The goal is to select one node from each frame to maximize the overall combined unary and binary weights. Given an arbitrary selection of nodes from the graph $s_{s} = \{s_s^i |_{i=1}^{F}\}$ ($F$ is the number of frames), the objective function is given by
\begin{equation}
\label{eqn_objective_function_single}
    \mathcal{M}_s(s_{s}) = \sum_{i=1}^{F}{\Phi_{s}(s_s^i)} + \lambda_s \cdot \sum_{i=1}^{F-1}{\Psi_{s}(s_s^i,s_s^{i+1})},
\end{equation}
where $\lambda_s$ is the parameter for adjusting the binary and unary weights, and $s_{s}^* = \arg \max_{s_{s}}(\mathcal{M}(s_{s}))$ gives the optimal solution. It is clear that the relational graph of this problem is a degenerated tree (i.e. single branch tree, please see Fig.\ref{fig_tracklet_hypothesis_graph}(a)), and as shown in Section \ref{sec_graphs}, the problem can be solved using dynamic programming efficiently. After the optimal solution is obtained, the selected nodes are removed from the graph and the next optimal solution is obtained. This process is iterated for multiple times in order to get several tracklets from the graph.

\subsection{Coupled Part Tracklets}
\label{sec_coupled_part_tracklets}

The relational graph for the coupled part tracklets generation is the same as for the single part; however, the nodes and edges are defined differently. In this case, each hypothesis node is composed of the locations of a pair of symmetric parts (e.g. left and right ankles). Fig.\ref{fig_tracklet_hypothesis_graph}(c) shows an illustration of the graph. Such design aims to model the symmetric relationship between coupled parts, including mutual location exclusions and appearance similarity in order to reduce double counting. As discovered in previous research \cite{Ramakrishna2013}, double counting is a key issue which severely hinders the pose estimation. Theoretically, tree based model \cite{Yang2011} lacks the ability to model spatial relationship of the coupled parts (e.g. left and right ankles). Furthermore, as discussed in Section \ref{sec_introduction}, attempting to model such spatial relationship would inevitably induce simple cycles in the graph which would severely increase the computational complexity. By introducing the coupled parts, this dilemma could be effectively solved. In the coupled part tracklet hypothesis graph, each node $r = (p,q)$ represents a composition of a pair of symmetric parts $p$ and $q$. Unary weights are assigned to the nodes, which represent the detection confidence and the compatibility between the two symmetric parts,  is defined as:
\begin{equation}
\label{eqn_coupled_part_unary_weight}
    \Phi_{c}(r) = \frac{(\Phi_{s}(r.p)+\Phi_{s}(r.q)) \cdot (\Lambda(r.p)^T \cdot \Lambda(r.q)))}{1+e^{- \left| r.p - r.q \right|/\theta}},
\end{equation}
where $\Phi_s$ is from Eqn.\ref{eqn_detection_score}, $r.p$ and $r.q$ respectively represent the left and right components of the coupled part $r$, $\Lambda(p)$ is the normalized color histogram of a local patch around $p$, the denominator is the inverse of a sigmoid function which penalizes the overlap of symmetric parts, and $\theta$ is the parameter that controls the penalty. The binary weights of the edges are computed as
\begin{equation}
\label{eqn_coupled_part_binary_weight}
    \Psi_{c}(r^f, r^{f+1}) = \Psi_{s}(r.p^f,r.p^{f+1}) + \Psi_{s}(r.q^f,r.q^{f+1}),
\end{equation}
where $\Psi_{s}$ is from Eqn. \ref{eqn_single_part_binary_weight}.

Similarly, the goal is to select one node (which is a composition of a pair of symmetric parts) from each frame to maximize the overall combined unary and binary weights. Given an arbitrary selection of nodes from the graph $s_{c} = \{s_c^i|_{i=1}^{F}\}$ (where $F$ is the number of frames), the objective function is
\begin{equation}
\label{eqn_objective_function_coupled}
    \mathcal{M}_c(s_{c}) = \sum_{i=1}^{F}{\Phi_{c}(s_c^i)} + \lambda_c \cdot \sum_{i=1}^{F-1}{\Psi_{c}(s_c^i,s_c^{i+1})},
\end{equation}
in which $\lambda_c$ is the parameter to adjust the binary and unary weights, and $s_c^* = \arg \max_{s_c}(\mathcal{M}(s_c))$ gives the optimal solution. As discussed in Section \ref{sec_graphs}, the problem can also be solved by dynamic programming efficiently, and iterated for multiple times to get several tracklets.

\begin{figure}
\begin{center}
\includegraphics[width=3.2in]{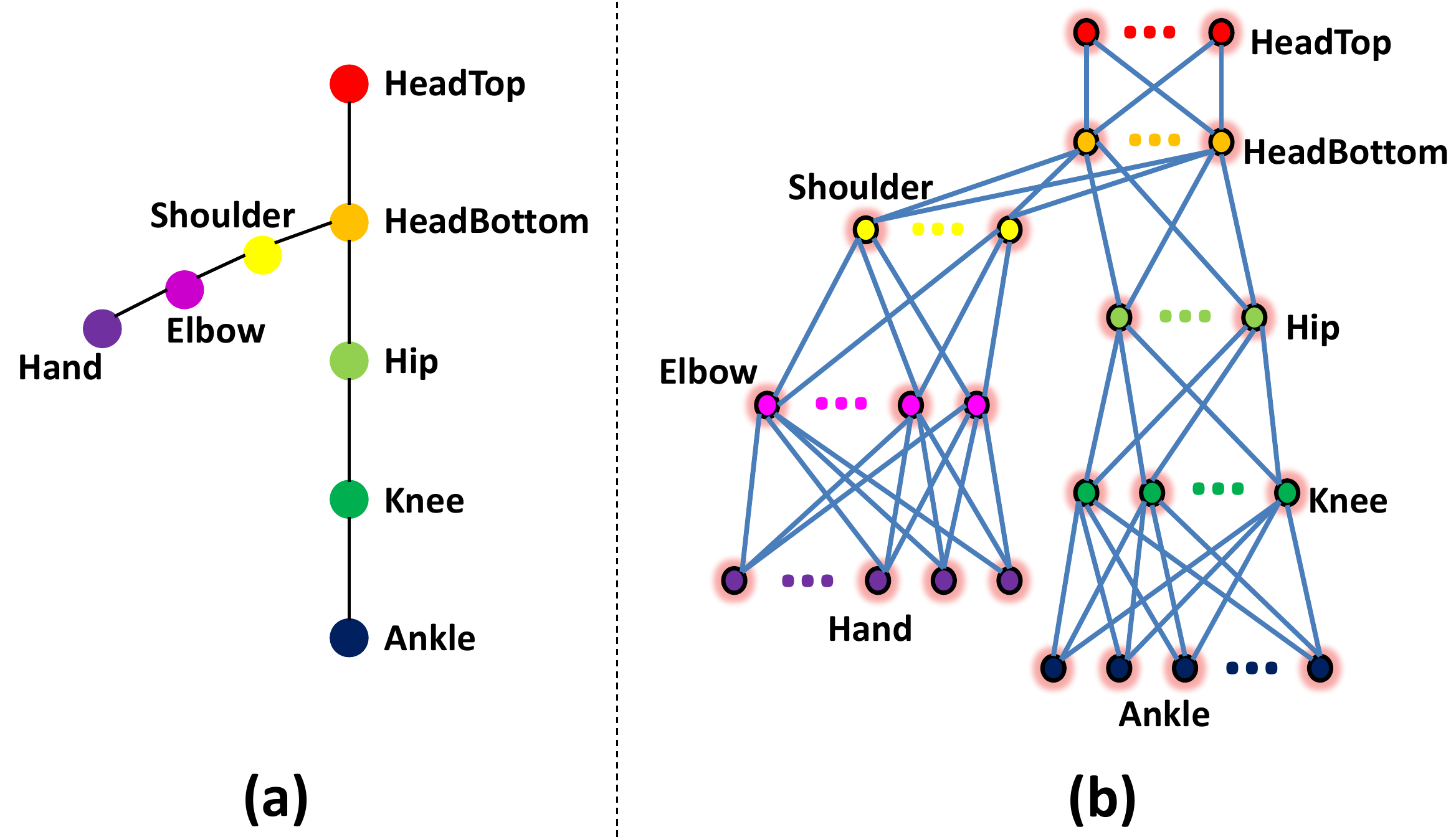}
\end{center}
   \caption{\textbf{Pose graphs}. (a) is the pose relational graph. Each node represents one abstract body part and edges represent the relationship between the connected body parts. (b) is the pose hypothesis graph. Each node is a tracklet for the part, and edges represent the spatial compatibility of connected nodes.}
   \label{fig_pose_hypothesis_graph}
\end{figure}

\subsection{Optimal Pose Estimation using Part Tracklets}
\label{sec_optimal_pose_estimation}

Since the best tracklets for each abstract body parts are obtained by the methods introduced in Section \ref{sec_single_part_tracklets} and \ref{sec_coupled_part_tracklets}, the next step is to select the best ones which are compatible. The relational graph, $G_T = (V_T,E_T)$, for this final tracklet based optimal pose estimation is shown in Fig.\ref{fig_pose_hypothesis_graph}(a). Each node represents an abstract body part, and the edges model the spatial relationships between them. Following the definitions of the abstract body parts and based on the part tracklets generated for these abstract body parts, a pose hypothesis graph can be built to get the optimal pose (as shown in Fig.\ref{fig_pose_hypothesis_graph}(b)). In this graph, each node represents an abstract body part tracklet and edges represent the spatial constraints. For each hypothesis tracklet node, $s$, depending on if it corresponds to a single part or a coupled part, $E_s(s)$ from Eqn.\ref{eqn_objective_function_single}, or  $E_c(s)$ from Eqn.\ref{eqn_objective_function_coupled} is used as its unary weight $\Phi_T(s)$. Let $\Psi_{d}(p_i,q_j) = \omega_{i,j} \cdot \psi(p_i-q_j)$ be the relative location score in \cite{Yang2011} ($\omega_{i,j}$ and $\psi$ are defined the same as in \cite{Yang2011}), the binary weight between a pair of adjacent single part tracklet nodes $s_{s} = \{s_s^i |_{i=1}^{F}\}$ and $t_{s} = \{t_s^i |_{i=1}^{F}\}$ is
\begin{equation}
    \Psi_T(s_s,t_s) = \sum_{i=1}^{F}{\Psi_{d}(s_s^i,t_s^i)},
\end{equation}
the binary weight between a single part tracklet node $s_{s} = \{s_s^i |_{i=1}^{F}\}$ and an adjacent coupled part tracklet node $t_{c} = \{t_c^i|_{i=1}^{F}\}$ is
\begin{equation}
    \Psi_T(s_s,t_c) = \sum_{i=1}^{F}{(\Psi_{d}(s_s^i,t_c^i.p) + \Psi_{d}(s_s^i,t_c^i.q))},
\end{equation}
and the binary weight between a pair of adjacent coupled tracklet part nodes $s_{c} = \{s_c^i|_{i=1}^{F}\}$ and $t_{c} = \{t_c^i|_{i=1}^{F}\}$ is
\begin{equation}
    \Psi_T(s_c,t_c) = \sum_{i=1}^{F}{(\Psi_{d}(s_c^i.p,t_c^i.p) + \Psi_{d}(s_c^i.q,t_c^i.q))}.
\end{equation}
The problem is to select only one tracklet for each abstract body part in order to maximize the combined unary (detection score) and binary (compatible score) weights. Given an arbitrary tree selected from the hypothesis graph $s_T = \{s_T^i|_{i=1}^{|V_T|}\}$, the objective function is
\begin{equation}
\label{eqn_objective_function_pose_tree}
    \mathcal{M}_T(s_T) = \sum_{v_T^i \in V_T}{\Phi_T(s_T^i)} + \lambda_T \cdot \sum_{(v_T^i,v_T^j) \in E_T}{\Psi_T(s_T^i,s_T^j)},
\end{equation}
where $\lambda_T$ is a parameter for adjusting the binary and unary weights, and as discussed in Section \ref{sec_graphs}, the optimal solution $s_T^* = \arg \max_{s_T}(\mathcal{M}(s_T))$ can also can be obtained by dynamic programming efficiently. The body part locations in each frame are extracted from this final optimal solution.

\subsection{Limb Alignment and Refinement}
\label{sec_limb_alignment_and_refinement}

\begin{figure*}
\begin{center}
\includegraphics[width=5in]{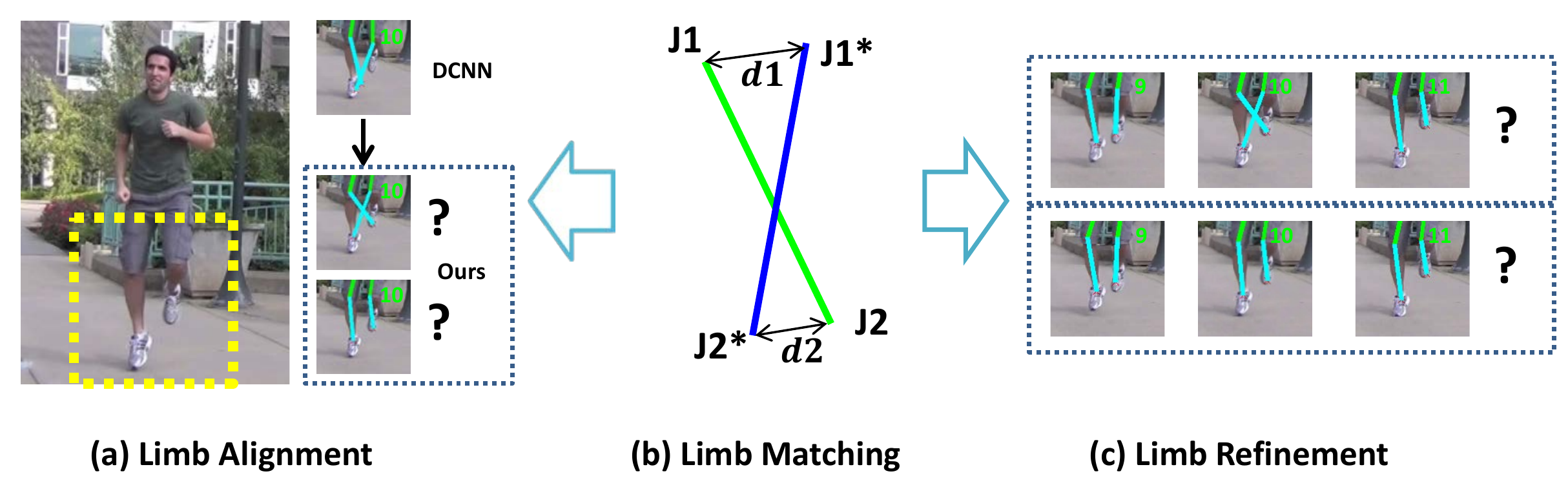}
   \caption{An Illustration for Limb Alignment and Refinement. (b) is an illustration for matching a single limb with its reference. $J1,J2$ are the two joints corresponding to the limb detection, and $J1^*,J2^*$ are the two joints for the reference limb. (a) shows the limb alignment process: by matching with the limb part configurations from DCNNGM results, the limb part configurations (lower legs) can be inferred correctly. (c) shows the limb refinement process: by matching with the limb configurations from neighboring frames, the wrong limb configuration (please see frame 10, lower legs) can be corrected. }
   \label{fig_limb_alignment_and_refinement}
\end{center}
\end{figure*}

\begin{figure*}
\begin{center}
\includegraphics[width=7in]{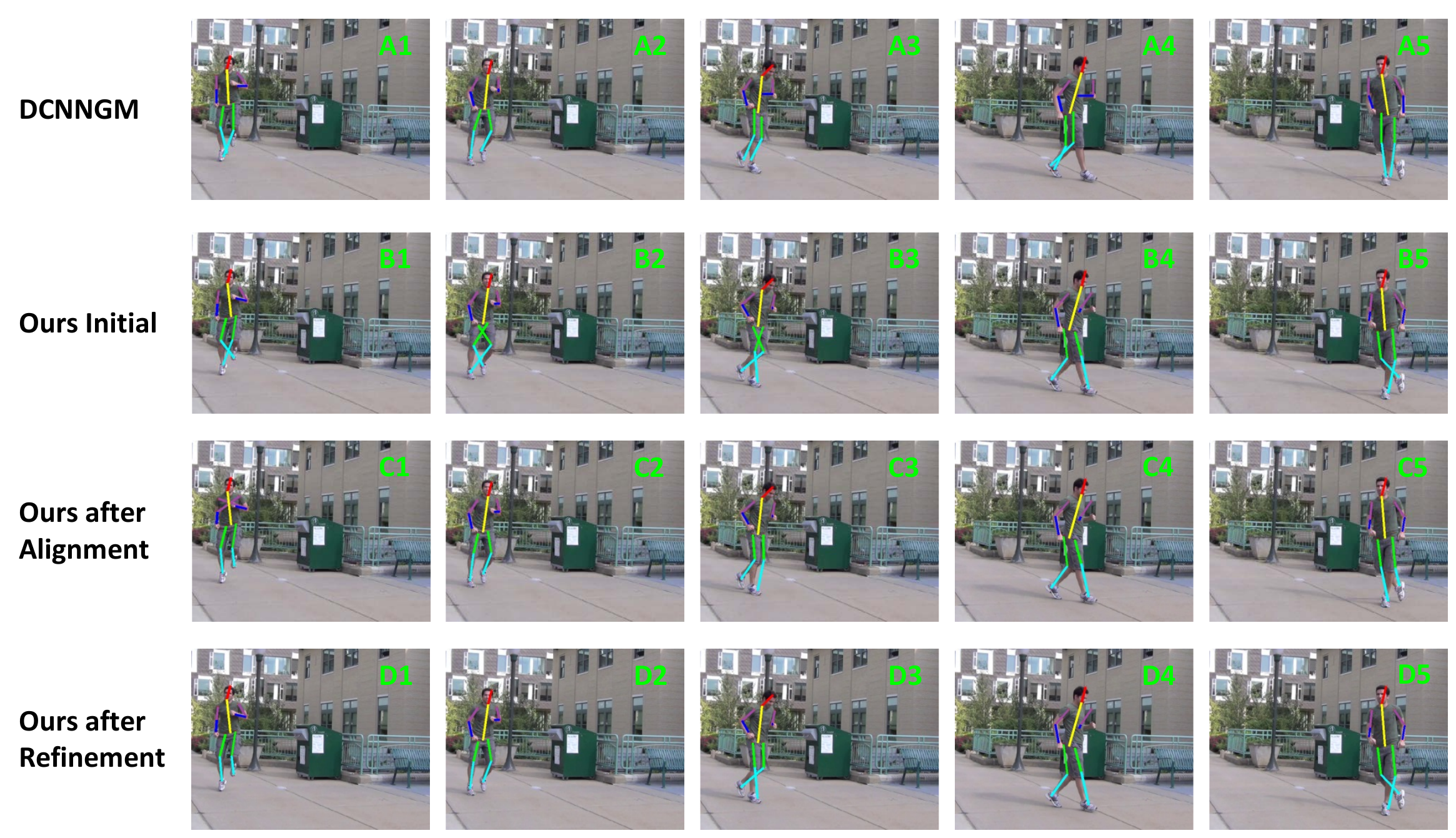}
   \caption{An Illustration for Limb Alignments and Refinements. ``DCNNGM" shows the initial pose estimation; ``Ours Initial" shows the results after ``Abstraction" and ``Association"; ``Ours after Alignment" and ``Ours after Refinement" show the results after these two steps respectively.}
   \label{fig_limb_alignment_and_refinement_results}
\end{center}
\end{figure*}

Most of the human pose estimation methods (e.g. \cite{Yang2011,Park2011,Chen2014}) do not infer the anatomical left/right body parts (some exceptions: \cite{Jain2014,Tompson2014}), which means that they only infer whether the parts are ``visually" left or right (we call this ``visual left/right" body parts). Of course, this relies on the training data annotations and theoretically there is nothing wrong with it; however, sometimes it is problematic. For instance, in Fig.\ref{fig_limb_alignment_and_refinement_results}(C5), the anatomical left and right knees appear ``visually" right and left respectively in the frame, but the anatomical left and right ankles appear ``visually" left and right respectively in the frame. In this case, failing to infer which knees correspond to which ankles will result in incorrect poses (e.g. Fig.\ref{fig_limb_alignment_and_refinement_results}(A3),(B3) and (C3)). The proposed framework, so far (until Section \ref{sec_optimal_pose_estimation}), does not have an effective mechanism to handle this issue, therefore we deal with it in this section.

Here, we define \textbf{limb inference problem} as the problem to determine the left-right relationship between adjacent coupled part pairs (e.g. to decide which knee matches with which ankle). Here ``limb" has a specific definition, which refers to ``upper arms", ``lower legs", and so on, not for ``full arms" or ``full legs".  So far, we do not have an effective scheme for limb inference, thus although each body part's  localization has improved, there are still some mismatches (please see Fig.\ref{fig_limb_alignment_and_refinement_results}). This is because  during the previous steps, we have not exploited limb inference results from image based pose estimation. We would like to utilize this information  from DCNNGM \cite{Chen2014} results. Although the body part localization results from DCNNGM are not as good as obtained by the proposed method, the limb inferences are much better and in most of the cases DCNNGM method can infer them correctly.

\subsubsection{Limb Alignment}

Based on the assumption that the limb inferences of the DCNNGM are correct in most of the video frames, we align our results with them. As illustrated in Fig.\ref{fig_limb_alignment_and_refinement} (b), a limb mismatching function is defined based on the joint locations of the reference limb and estimated limb. The mismatch score is defined as:
\begin{equation}
    \label{eqn_limb_mismatching_single_limb}
    \mathcal{M}_{s}(L,\mathcal{L}) = (\| L(1)-\mathcal{L}(1) \|^2_2 + \| L(2)-\mathcal{L}(2) \|^2_2),
\end{equation}
where $L$ and $\mathcal{L}$ denote the detected limb and reference limb respectively (in this situation, $\mathcal{L}$ is the results from DCNNGM \cite{Chen2014}). $L(1),L(2),\mathcal{L}(1),\mathcal{L}(2)$ are the two joint locations of the limbs.

For a pair of limbs (i.e. the left and right limbs), the mismatching score is defined as
\begin{multline}
    \label{eqn_limb_mismatching_single_frame}
    \mathcal{M}_{p}(L^l,L^r|\mathcal{L}^l,\mathcal{L}^r) = \mathcal{C}(L^l,L^r|\mathcal{L}^l,\mathcal{L}^r) \\ \cdot (\mathcal{M}_{s}(L^l,\mathcal{L}^l) + \mathcal{M}_{s}(L^r,\mathcal{L}^r)),
\end{multline}
where $\mathcal{C}(L^l,L^r|\mathcal{L}^l,\mathcal{L}^r)$ is a limb intersection penalty function which is defined as
\begin{equation}
    \label{eqn_intersection_penalty}
    \mathcal{C}(L^l,L^r|\mathcal{L}^l,\mathcal{L}^r) =
    \begin{cases}
        c & \text{$L^l,L^r$ are compatible with $\mathcal{L}^l,\mathcal{L}^r$} \\
        1-c & \text{otherwise}
    \end{cases}
\end{equation}
where ``compatible'' means both $L^l,L^r$ and $\mathcal{L}^l,\mathcal{L}^r$ intersect each other, or both of them do not intersect each other.

Using the mismatching function for each pair of limbs in each frame, the configuration which has lower mismatching score is selected to be the correct limb inference. For instance in Fig.\ref{fig_limb_alignment_and_refinement}(a), compared to the DCNNGM reference, the bottom configuration of the lower legs has lower mismatching score, thus it is considered to be the correct limb inference for lower legs. It can be seen from this example that the proposed method gives better joint localization results (i.e. for feet), however potentially wrong limb inferences. Using DCNNGM limb inference results, the aligned results are now better in both the joint localization and limb inferences.

\subsubsection{Limb Refinement}

The limb alignment can correct most of the mismatched limbs; however, sometimes DCNNGM also makes wrong limb inference, therefore only relying on the limb alignment does not solve the limb inference problem completely. Consequently, the limb inferences in neighboring frames is utilized to refine the results. Similar to Eqn.\ref{eqn_limb_mismatching_single_frame}, a mismatching score is introduced for two neighboring frames (frame $f$ and $f+1$):
\begin{multline}
    \label{eqn_limb_mismatching_neighboring_frames}
    \mathcal{M}_{n}(L^l_{f},L^l_{f+1},L^r_{f},L^r_{f+1}|\mathcal{L}^l_{f},\mathcal{L}^l_{f+1},\mathcal{L}^r_{f},\mathcal{L}^r_{f+1})
    \\
    = \mathcal{C}(L^l_{f},L^r_{f}|\mathcal{L}^l_{f},\mathcal{L}^r_{f}) \cdot \mathcal{C}(L^l_{f+1},L^r_{f+1}|\mathcal{L}^l_{f+1},\mathcal{L}^r_{f+1}) \\
    \cdot (\mathcal{M}_{s}(L^l_{f},L^l_{f+1}) + \mathcal{M}_{s}(L^r_{f},L^r_{f+1})),
\end{multline}
where $\mathcal{M}_s$ and $\mathcal{C}$ are defined similarly as in Eqn.\ref{eqn_limb_mismatching_single_limb} and Eqn.\ref{eqn_intersection_penalty}. The best configurations of the limbs through all the video frames are obtained simultaneously by minimizing the sum of mismatching scores of all consecutive frames in the video using simple dynamic programming (i.e. similar to Eqn.\ref{eqn_objective_function_tree}, but without unary terms).

Fig. \ref{fig_limb_alignment_and_refinement_results} shows some results of the proposed limb alignment and refinement schemes. As can be seen from the results, joints are much better  localized  (please see the feet in Fig.\ref{fig_limb_alignment_and_refinement_results}(B4) and (B5) compared to (A4) and (A5)); however, the limb inferences are not correct in some frames (e.g. the lower legs in Fig.\ref{fig_limb_alignment_and_refinement_results}(B1) and (B2)). After the limb alignment, most of the limb inferences are corrected (e.g. the lower legs in Fig.\ref{fig_limb_alignment_and_refinement_results}(B1) and (B2)), however some of them are still not perfect (e.g. Fig.\ref{fig_limb_alignment_and_refinement_results}(C3) and (C5)). After the limb refinement, these errors are corrected (e.g. Fig.\ref{fig_limb_alignment_and_refinement_results}(C3) and (C5)). Please note that the ``limb alignment" sometimes deteriorate the results (e.g. please compare Fig.\ref{fig_limb_alignment_and_refinement_results}(C5) to (B5)), and this is because the DCNNGM made a wrong limb inference (e.g. in Fig.\ref{fig_limb_alignment_and_refinement_results}(A5)), which failed the limb alignment process. Nevertheless, the ``limb refinement" process corrected these errors and we achieve better performance (e.g. Fig.\ref{fig_limb_alignment_and_refinement_results}(D3) and (D5)).

\section{Experiments}
\label{sec_experiments}

\subsection{Datasets}

We evaluated our method on three publicly available datasets:

\textbf{Outdoor Pose Dataset}: this dataset was collected by the authors of \cite{Ramakrishna2013}, which contains six video sequences from outdoor scenes. There are a lot of self-occlusions of the body parts in this dataset.  Annotations of more than 1,000 frames are provided by the authors.

\textbf{Human Eva-I}: this dataset \cite{Sigal2010} contains human activities in indoor controlled conditions. The activities are synchronized with a ground truth of 3D motion capture data, which can be converted into 2D joint locations. In order to have a fair comparison with  \cite{Ramakrishna2013}, we use 250 frames from the sequences: \emph{S1\_Walking}, \emph{S1\_Jog}, \emph{S2\_Jog} captured by camera 1.

\textbf{N-Best Dataset}, this dataset was collected by the authors of \cite{Park2011} and has four sequences in total. As a fair comparison to \cite{Ramakrishna2013}, we also report results on sequences \emph{walkstraight} and \emph{baseball}.

\subsection{Evaluation Metrics}

Similar to \cite{Ramakrishna2013}, we use PCP and KLE to evaluate our results. Probability of a Correct Pose (PCP) \cite{Ferrari2008} is a standard evaluation metric, which measures the percentage of correctly localized body parts within a threshold. Keypoint Localization  Error (KLE) \cite{Ramakrishna2013} measures the average Euclidean distance from the ground truth to the estimated keypoints, normalized by the size of the head in each frame.

\begin{table*}
\caption{Comparisons with the state-of-the-art methods and intermediate stages of the proposed method, on three publicly available datasets. Note that PCP is an accuracy measure, so the larger the better, with a max of $1$; and KLE is an error measure, so the smaller the better. Please refer to Section \ref{sec_comparison_results} for a detailed discussion.}
\begin{center}
\begin{tabular}{ |c|c|c|c|c|c|c|c|c|  }
 \hline
 \rowcolor{lightgray} \multicolumn{9}{|c|}{\bf Outdoor Dataset \cite{Ramakrishna2013}} \\
 \hline
 {\bf Metric } & {\bf Method } &{\bf Head }  & {\bf Torso } & {\bf U.L } & {\bf L.L} & {\bf U.A. } & {\bf L.A.} & {\bf Average}\\
 \hline
 \multirow{3}{2em}{PCP}     & Sym-Trk \cite{Ramakrishna2013}  & 0.99          & 0.86          & 0.95          & 0.96          & 0.86          & 0.52 & 0.86
 \\                         & N-Best \cite{Park2011}         & 0.99          & 0.83          & 0.92          & 0.86          & 0.79          & 0.52 & 0.82
 \\                         & Mix-Part \cite{Cherian2014}      & 0.87          & 0.97          & 0.68          & 0.89          & 0.78          & 0.52 & 0.79
 \\                         & DCNNGM \cite{Chen2014}    & 1.00          & 1.00          & 0.98          & 0.94          & 0.94          & 0.85 & 0.95
 \\                         & Ours HPEV \cite{Zhang2015} (Baseline)          & 0.92          & 1.00          & 0.84          & 0.73          & 0.68          & 0.47 & 0.77
 \\                         & Ours HPEV \cite{Zhang2015} (Abt. Only)         & 0.99          & 1.00          & 0.89          & 0.77          & 0.72          & 0.53 & 0.82
 \\                         & Ours HPEV \cite{Zhang2015} (Asc. Only)         & 0.99          & 1.00          & 0.87          & 0.76          & 0.79          & 0.56 & 0.83
 \\                         & Ours HPEV \cite{Zhang2015}                   & 0.99          & 1.00          & \textbf{1.00} & 0.97          & 0.91          & 0.66 & 0.92
 \\                         & Ours (HPEV \cite{Zhang2015} + DCNNGM)        & 1.00          & 1.00          & 0.96          & 0.93          & 0.81          & 0.77 & 0.91
 \\                         & Ours Final (HPEV \cite{Zhang2015} + DCNNGM + A.R.) & 1.00      & 1.00          & 0.97          & \textbf{0.98} & \textbf{0.95} & \textbf{0.88} & \textbf{0.96}
 \\
 \hline
 \multirow{3}{2em}{KLE}     & Sym-Trk\cite{Ramakrishna2013}  & 0.39          & 0.58          & 0.48          & 0.48          & 0.88          & 1.42 & 0.71
 \\                         & N-Best \cite{Park2011}         & 0.44          & 0.58          & 0.55          & 0.69          & 1.03          & 1.65 & 0.82
 \\                         & Mix-Part \cite{Cherian2014}      & 0.31          & 0.72          & 0.91          & 0.36          & 0.44          & 0.72 & 0.58
  \\                         & DCNNGM \cite{Chen2014}   & 0.18          & 0.15          & \textbf{0.28} & 0.32          & 0.27          & 0.34 & 0.26
 \\                         & Ours HPEV \cite{Zhang2015} (Baseline)          & 0.58          & 0.45          & 0.61          & 0.78          & 0.75          & 1.11 & 0.71
 \\                         & Ours HPEV  \cite{Zhang2015} (Abt. Only)         & \textbf{0.16} & 0.23          & 0.48          & 0.69          & 0.55          & 0.78 & 0.48
 \\                         & Ours HPEV \cite{Zhang2015} (Asc. Only)         & \textbf{0.16} & 0.20          & 0.47          & 0.64          & 0.44          & 0.71 & 0.44
 \\                         & Ours HPEV \cite{Zhang2015}                    & 0.19          & 0.22          & 0.35          & 0.37          & 0.41          & 0.61 & 0.36
 \\                         & Ours (HPEV \cite{Zhang2015} + DCNNGM)        & 0.18          & 0.15          & 0.38          & 0.36          & 0.38          & 0.38 & 0.30
 \\                         & Ours Final (HPEV \cite{Zhang2015} + DCNNGM + A.R.) & 0.18      & 0.15          & \textbf{0.28} & \textbf{0.26} & \textbf{0.25} & \textbf{0.30} & \textbf{0.24}
 \\
 \hline
 \rowcolor{lightgray} \multicolumn{9}{|c|}{\bf Human Eva-I Dataset \cite{Sigal2010}} \\
 \hline
 {\bf Metric } & {\bf Method } &{\bf Head }  & {\bf Torso } & {\bf U.L } & {\bf L.L} & {\bf U.A. } & {\bf L.A.} & {\bf Average}\\
 \hline
 \multirow{3}{2em}{PCP}     & Sym-Trk\cite{Ramakrishna2013}  & 0.99          & 1.00          & 0.99          & \textbf{0.98} & \textbf{0.99} & 0.53 & 0.91
 \\                         & N-Best \cite{Park2011}         & 0.97          & 0.97          & 0.97          & 0.90          & 0.83          & 0.48 & 0.85
 \\                         & Mix-Part \cite{Cherian2014}      & 0.99          & 1.00          & 0.90          & 0.89          & 0.96          & 0.62 & 0.89
 \\                         & DCNNGM \cite{Chen2014}    & 1.00          & 1.00          & 1.00          & 0.93          & 0.98          & 0.74 & 0.94
 \\                         & Ours HPEV \cite{Zhang2015} (Baseline)          & 1.00          & 1.00          & 0.93          & 0.62          & 0.44          & 0.24 & 0.71
 \\                         & Ours HPEV \cite{Zhang2015} (Abt. Only)         & 1.00          & 1.00          & 0.98          & 0.66          & 0.43          & 0.30 & 0.73
 \\                         & Ours HPEV \cite{Zhang2015} (Asc. Only)         & 1.00          & 1.00          & 0.94          & 0.62          & 0.45          & 0.27 & 0.71
 \\                         & Ours HPEV \cite{Zhang2015}                    & 1.00          & 1.00          & 1.00          & 0.94          & 0.93          & 0.67 & 0.92
 \\                         & Ours (HPEV \cite{Zhang2015} + DCNNGM)        & 1.00          & 1.00          & 1.00          & 0.96          & 0.63          & 0.67 & 0.88
 \\                         & Ours Final (HPEV \cite{Zhang2015} + DCNNGM + A.R.) & 1.00      & 1.00          & 1.00          & 0.96          & 0.97          & \textbf{0.78} & \textbf{0.95}
 \\
 \hline
 \multirow{3}{2em}{KLE}     & Sym-Trk \cite{Ramakrishna2013}  & 0.27          & 0.48          & \textbf{0.13} & 0.22          & 1.14          & 1.07 & 0.55
 \\                         & N-Best \cite{Park2011}         & 0.23          & 0.52          & 0.24          & 0.35          & 1.10          & 1.18 & 0.60
 \\                         & Mix-Part \cite{Cherian2014}      & \textbf{0.13}& 0.40  & 0.23          & 0.16          & \textbf{0.14} & 0.24 & 0.22
 \\                         & DCNNGM \cite{Chen2014}    & 0.19          & 0.42          & 0.15          & 0.17          & 0.17          & 0.22 & 0.22
 \\                         & Ours HPEV \cite{Zhang2015} (Baseline)          & 0.17          & 0.40          & 0.34          & 0.45          & 0.66          & 0.84 & 0.48
 \\                         & Ours HPEV \cite{Zhang2015} (Abt. Only)         & 0.17          & 0.41          & 0.29          & 0.41          & 0.66          & 0.75 & 0.45
 \\                         & Ours HPEV \cite{Zhang2015} (Asc. Only)         & 0.17          & 0.39          & 0.33          & 0.42          & 0.63          & 0.74 & 0.45
 \\                         & Ours HPEV \cite{Zhang2015}                   & 0.16          & 0.42          & \textbf{0.13} & 0.15          & 0.20          & 0.24 & 0.22
 \\                         & Ours (HPEV \cite{Zhang2015} + DCNNGM)        & 0.19          & 0.42          & 0.17          & 0.17          & 0.29          & 0.26 & 0.25
 \\                         & Ours Final (HPEV \cite{Zhang2015} + DCNNGM + A.R.) & 0.19      & 0.42          & 0.14          & \textbf{0.14}          & 0.17          & \textbf{0.20} & \textbf{0.21}
 \\
 \hline
 \rowcolor{lightgray} \multicolumn{9}{|c|}{\bf N-Best Dataset \cite{Park2011}} \\
 \hline
{\bf Metric } & {\bf Method } &{\bf Head }  & {\bf Torso } & {\bf U.L } & {\bf L.L} & {\bf U.A. } & {\bf L.A.} & {\bf Average}\\
 \hline
 \multirow{3}{2em}{PCP}     & Sym-Trk \cite{Ramakrishna2013}  & 1.00          & 0.69          & 0.91          & 0.89          & 0.85          & 0.42 & 0.80
 \\                         & N-Best \cite{Park2011}         & 1.00          & 0.61          & 0.86          & 0.84          & 0.66          & 0.41 & 0.73
 \\                         & Mix-Part \cite{Cherian2014}      & 1.00          & 1.00          & 0.91          & 0.90          & 0.69          & 0.39 & 0.82
 \\                         & DCNNGM \cite{Chen2014}    & 1.00          & 1.00          & 0.91          & 0.91          & 0.96          & 0.77 & 0.92
 \\                         & Ours HPEV \cite{Zhang2015} (Baseline)          & 1.00          & 1.00          & 0.92          & 0.87          & 0.87          & 0.52 & 0.86
 \\                         & Ours HPEV \cite{Zhang2015} (Abt. Only)         & 1.00          & 1.00          & 0.91          & 0.89          & 0.87          & 0.65 & 0.89
 \\                         & Ours HPEV \cite{Zhang2015} (Asc. Only)         & 1.00          & 1.00          & \textbf{0.93} & 0.91          & 0.87          & 0.55 & 0.88
 \\                         & Ours HPEV \cite{Zhang2015}                    & 1.00          & 1.00          & 0.92          & \textbf{0.94} & 0.93          & 0.65 & 0.91
 \\                         & Ours (HPEV \cite{Zhang2015} + DCNNGM)        & 1.00          & 1.00          & 0.89          & 0.87          & 0.80          & 0.78 & 0.89
 \\                         & Ours Final (HPEV \cite{Zhang2015} + DCNNGM + A.R.) & 1.00      & 1.00          & 0.92          & 0.92          & \textbf{0.97} & \textbf{0.87} & \textbf{0.95}
 \\
 \hline
 \multirow{3}{2em}{KLE}     & Sym-Trk \cite{Ramakrishna2013}  & 0.53          & 0.88          & 0.67          & 1.01          & 1.70          & 2.68 & 1.25
 \\                         & N-Best \cite{Park2011}         & 0.54          & 0.74          & 0.80          & 1.39          & 2.39          & 4.08 & 1.66
 \\                         & Mix-Part \cite{Cherian2014}      & 0.15          & 0.23          & 0.31          & 0.37          & 0.46          & 1.18 & 0.45
 \\                         & DCNNGM \cite{Chen2014}    & 0.14          & 0.15          & 0.29          & 0.40          & 0.23          & 0.43 & 0.27
 \\                         & Ours HPEV \cite{Zhang2015} (Baseline)          & 0.15          & 0.19          & 0.36          & 0.49          & 0.32          & 0.84 & 0.39
 \\                         & Ours HPEV \cite{Zhang2015} (Abt. Only)         & 0.15          & 0.19          & 0.31          & 0.43          & 0.34          & 0.60 & 0.34
 \\                         & Ours HPEV \cite{Zhang2015} (Asc. Only)         & 0.15          & 0.17          & 0.27          & 0.42          & 0.29          & 0.68 & 0.33
 \\                         & Ours HPEV \cite{Zhang2015}                    & 0.15          & 0.17          & \textbf{0.24} & 0.37          & 0.30          & 0.60 & 0.31
 \\                         & Ours (HPEV \cite{Zhang2015} + DCNNGM)        & 0.14          & 0.15          & 0.34          & 0.36          & 0.36          & 0.46 & 0.30
 \\                         & Ours Final (HPEV \cite{Zhang2015} + DCNNGM + A.R.) & 0.14      & 0.15          & 0.26          & \textbf{0.29} & \textbf{0.23} & \textbf{0.34} & \textbf{0.24}
 \\
 \hline
\end{tabular}
\end{center}
\label{table_results}
\end{table*}

\begin{figure*}
\begin{center}
\includegraphics[width=7in]{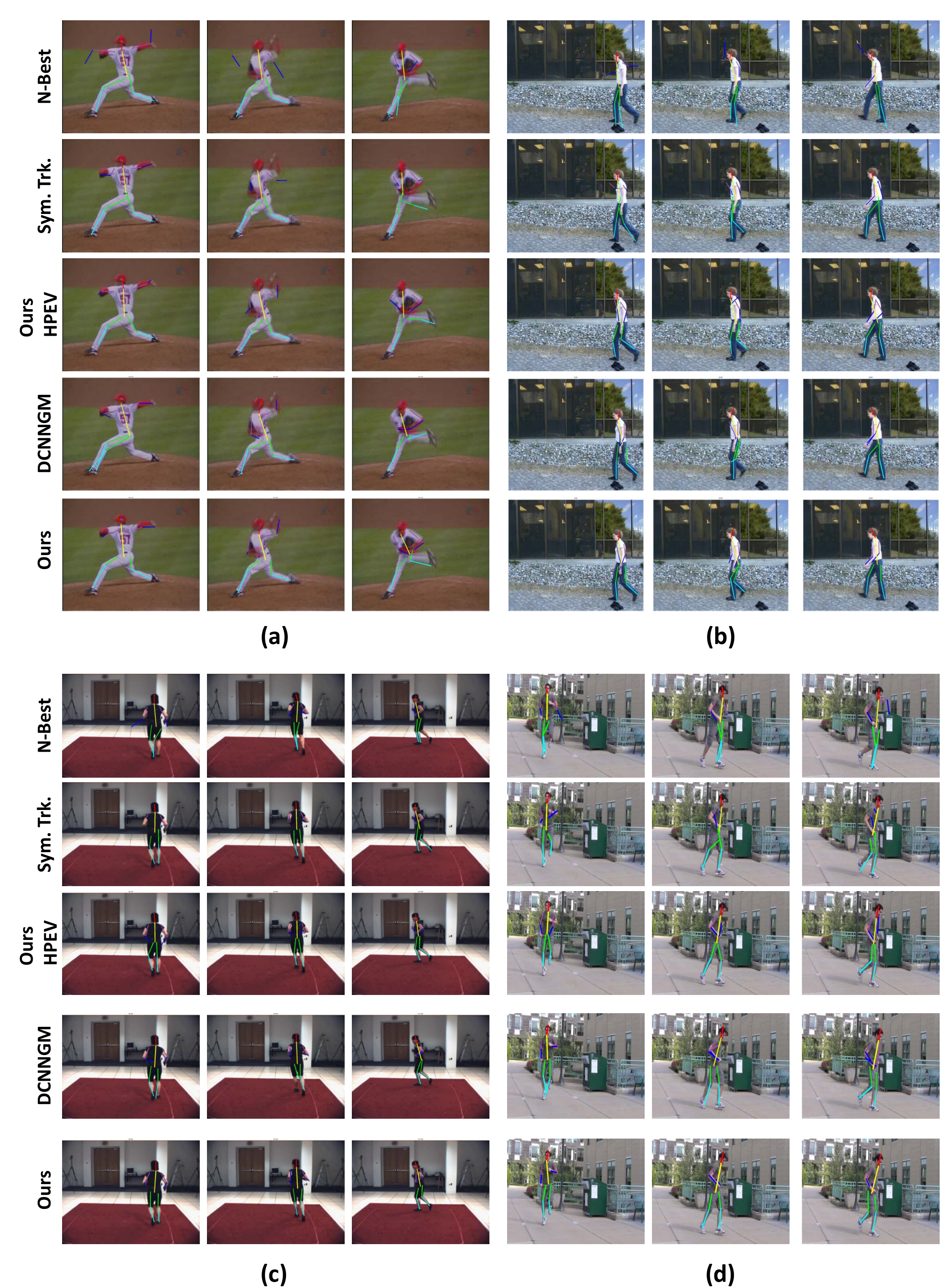}
\end{center}
   \caption{Examples and Comparisons with the State-of-the-art Methods: N-Best \cite{Park2011}, Symmetric Tracking (Sym. Trk.) \cite{Ramakrishna2013} , HPEV \cite{Zhang2015}, DCNNGM \cite{Chen2014} and ours. (a) and (b) are from N-Best Dataset; (c) is from Human Eva I dataset; and (d) is from Ourdoor Pose Dataset. Body parts are shown in different colors.}
   \label{fig_results1}
\end{figure*}

\begin{figure*}
\begin{center}
\includegraphics[width=7in]{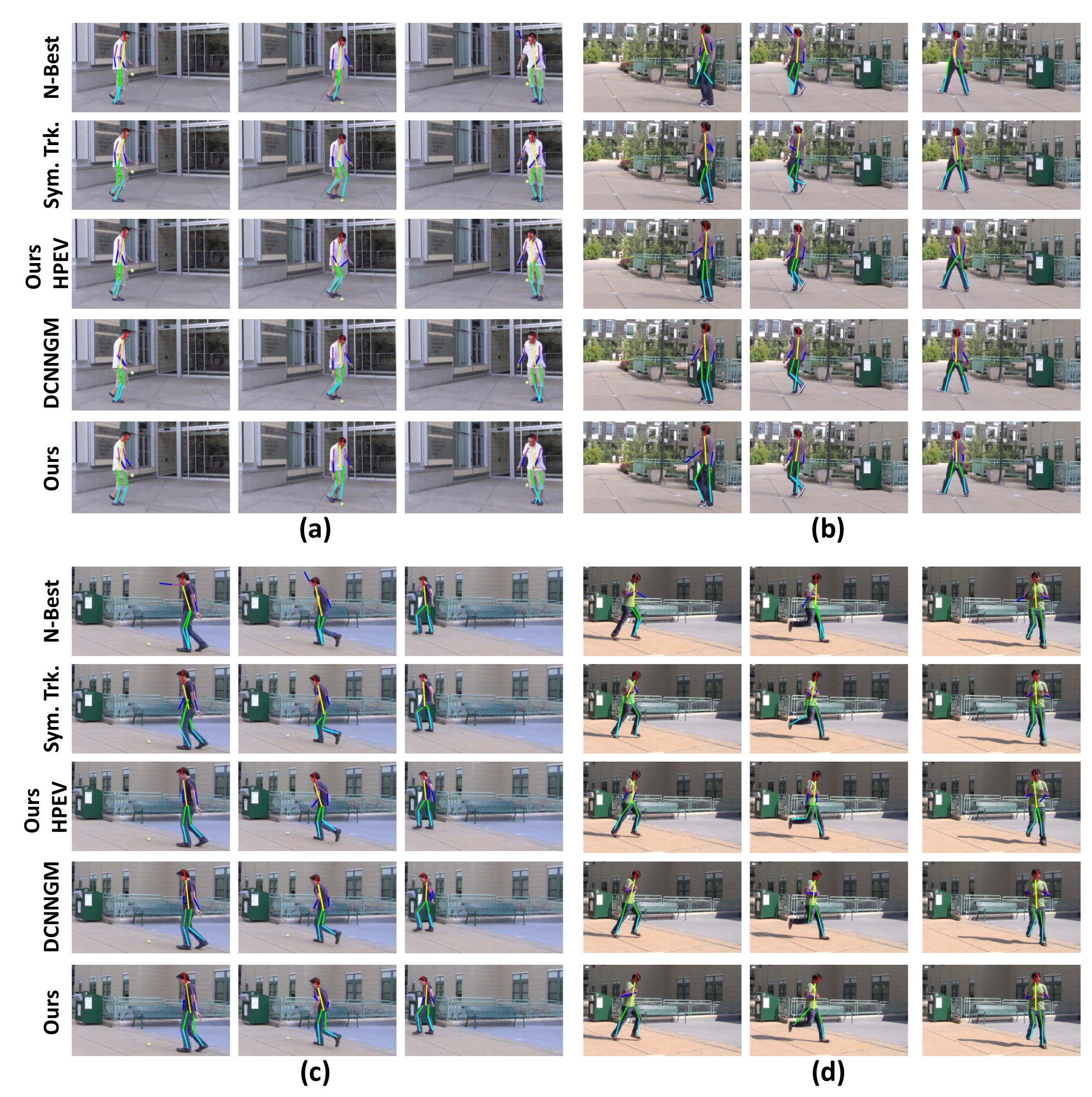}
\end{center}
   \caption{Examples and Comparisons with the State-of-the-art Methods: N-Best \cite{Park2011}, Symmetric Tracking (Sym. Trk.) \cite{Ramakrishna2013} , HPEV \cite{Zhang2015}, DCNNGM \cite{Chen2014} and ours. These results are from Ourdoor Pose Dataset. Body parts are shown in different colors.}
   \label{fig_results2}
\end{figure*}

\begin{figure*}
\begin{center}
\includegraphics[width=7in]{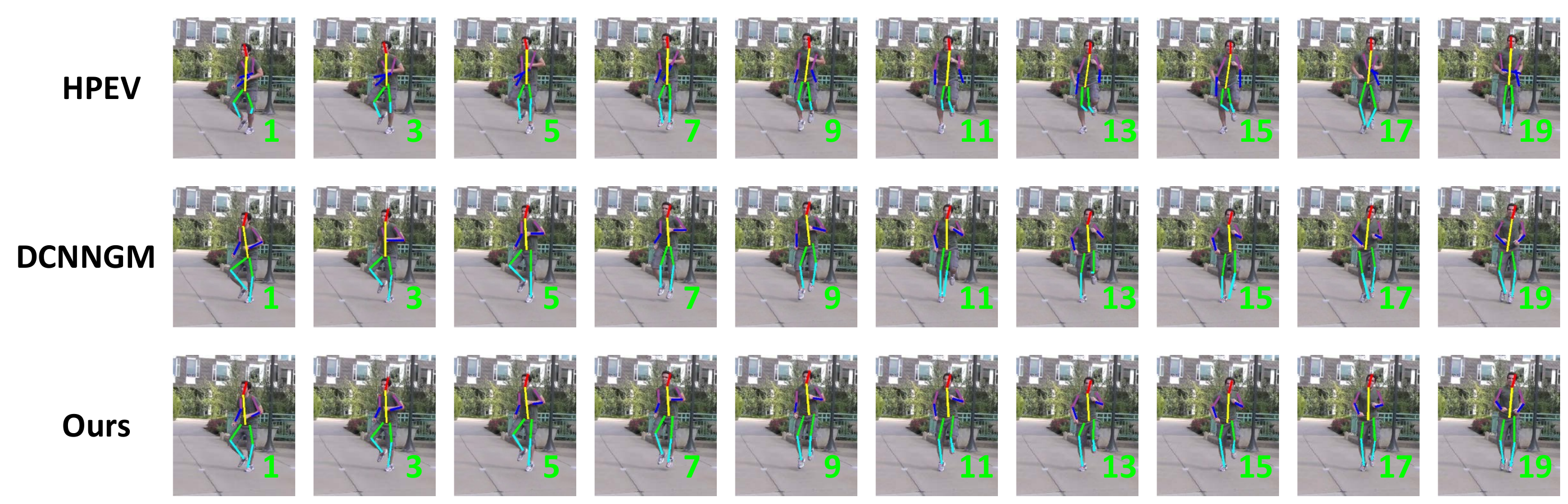}
\end{center}
   \caption{Detailed Comparisons of Results from a Video Sequence. Here we want to show the consistency of the results for the proposed method. Please note the pose estimations for the legs: the proposed method can localize the two ankles accurately and smoothly, while ``HPEV" results are also smooth but cannot localize the feet correctly in several frames, and ``DCNNGM" results are quite noisy and the performance vary a lot through the frames.}
   \label{fig_results_detail}
\end{figure*}

\subsection{Comparison Results}
\label{sec_comparison_results}

We compare the proposed method with three state-of-the-art video based human pose estimation methods: N-Best method \cite{Park2011}, Symmetric Tracking method \cite{Ramakrishna2013}, Mixing Body-part method \cite{Cherian2014}, and a deep learning baseline method (i.e. DCNNGM \cite{Chen2014}); we did not compare with some upper body pose estimation/tracking methods \cite{Sapp2011,Tokola2013}, since they focus on the modeling of hands/elbows using motion and appearance features but do not handle other body parts. Since \cite{Cherian2014} was designed for upper-body pose estimation, we re-implemented its algorithm by reusing most of their implementation but extended it to a full-body detection model. Quantitative results are shown in Table \ref{table_results}, and qualitative results are shown in Fig.\ref{fig_results1} and Fig.\ref{fig_results2}. Note that the figures for Symmetric Tracking method are re-produced from figures in \cite{Ramakrishna2013} since the code is not publicly available.

We also show detailed results to analyze the contributions of each step of the proposed method. In Table \ref{table_results}, ``Ours HPEV (Baseline)" is the proposed method without using ``Abstraction" or ``Association"; ``Ours HPEV (Abt. Only)" shows the results for only applying the ``Abstraction" step of our method; ``Ours HPEV (Ass. Only)" shows the results for only using the ``Association" step of the proposed method; and ``Ours HPEV" is the proposed method with ``Abstraction" and ``Association". From these results we found that ``Abstraction" is more important than ``Association" in the proposed method, due to the fact that it contributes more to the quantitative improvement. Furthermore, ``Ours (HPEV + DCNNGM)" is the proposed method with DCNNGM \cite{Chen2014} detection + ``Abstraction" + ``Association"; and ``Ours Final (HPEV + DCNNGM + A.R.)" shows the results with ``Limb Alignment and Refinement" steps added to the system. As we can see from the results, combining DCNNGM with our method does not necessarily improve the results (actually results deteriorate slightly). This is due to the fact that   our original method \cite{Zhang2015} can not infer the limb configurations accurately (please see Fig.\ref{fig_limb_alignment_and_refinement_results}); however, the proposed method have better joint localization results (please see Fig.\ref{fig_limb_alignment_and_refinement_results}). Although PCP and KLE do not measure the joint localization accuracy,  we can see the joint localization improvement from the figure and also from the final results. Our method boosts the average PCP accuracy above $0.95$ for all the datasets. Compared to the state-of-the-art methods,the average improvement in KLE is more than $10\%$, for PCP we reduced the error by about $15\%$, which means the proposed method can localize the body parts much more accurately.

\begin{figure}
\begin{center}
\includegraphics[width=3.2in]{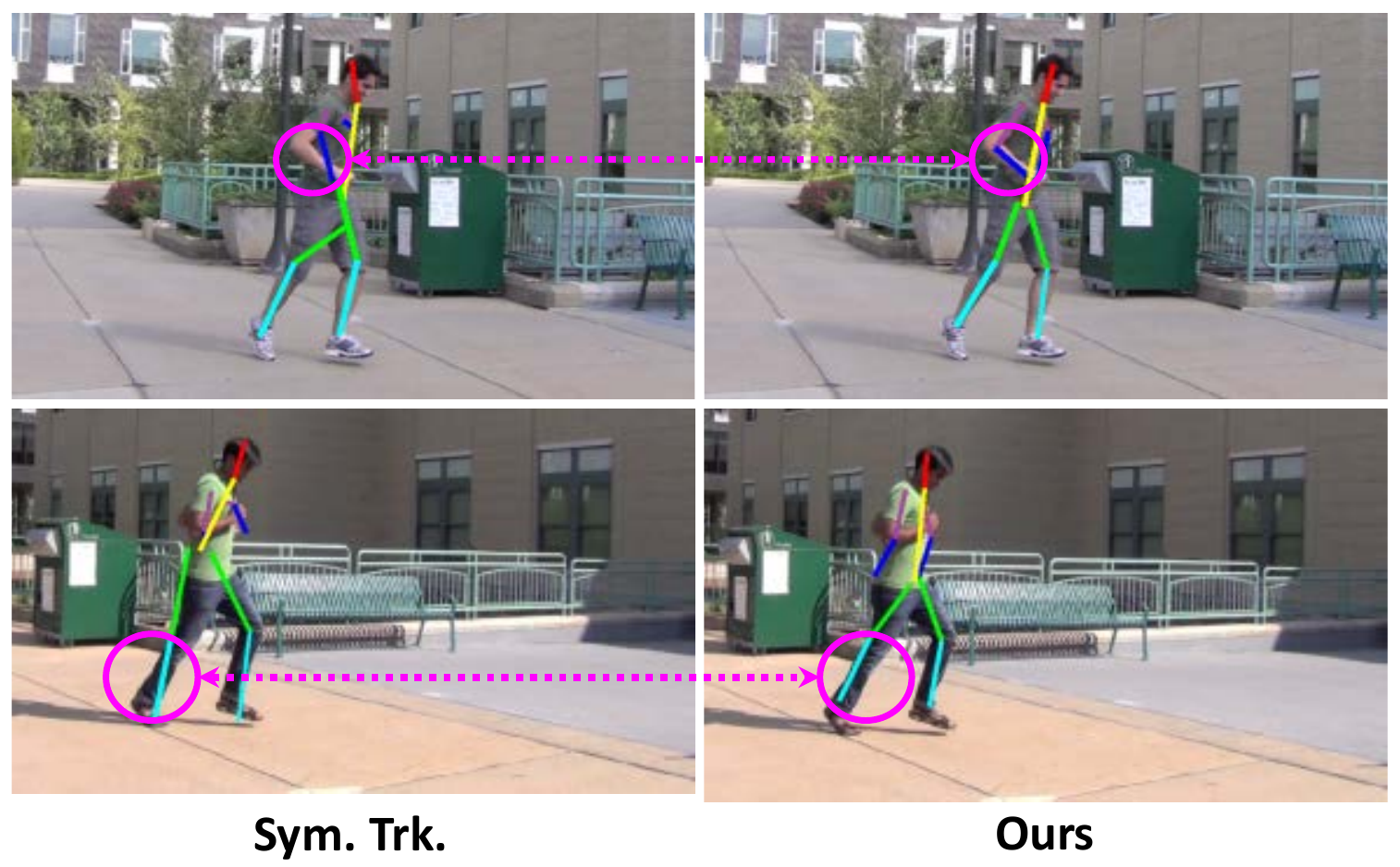}
\end{center}
   \caption{The Detailed Comparisons of the Results Obtained by Symmetric Tracking (Sym. Trk. \cite{Ramakrishna2013}) and the Proposed Method. This example shows that the body part location estimation could be considered `correctly' by PCP \cite{Ferrari2008}, but would induce larger error using KLE \cite{Ramakrishna2013}. For example, for the arms in the top figures and legs in the bottom figures, results for both methods satisfy the PCP criteria; however, our method would have much smaller KLE since the estimated body parts are much closer to the ground truth.
   }
   \label{fig_result_example}
\end{figure}

From these results,  a good question to ask is  why there is more improvement in KLE compared to PCP. We believe PCP is a relatively loose measure, and the state-of-the-art methods can already get about $0.8$, which means 80 percent of the parts could be localized correctly following the PCP criteria. The possible improvement margin is only $0.2$ and our improvement is $0.15$. Another reason is that, based on the definition of PCP \cite{Ferrari2008}, the parts would be considered `correctly' localized if the overlap ratio is below a threshold; however, KLE measures the distance between the body part and the ground truth. Therefore, sometimes it happens that a body part is considered `correctly' localized by PCP but still it is relatively far from the ground truth. As shown in Fig.\ref{fig_result_example}, both body parts shown in pink circles are `correctly' localized following the PCP criteria, however it is clear that our part estimations are much closer to the real locations of the body parts.

Sometimes the results are not fully reflected by the numbers. We show detailed results for some consecutive frames (sampled every two frames) in Fig.\ref{fig_results_detail}, and here we only compare our method with two best performing methods. As we can see clearly that the proposed method performs much better (please note the legs, in particular), and the poses are much consistent and all the body parts are located quite precisely. However, once we look at the numbers, the PCP results are exactly the same, and KLE numbers do not reflect the big improvement.

\textbf{Implementation Details:} We process 15 consecutive frames each time. For Eqn.\ref{eqn_objective_function_single} and \ref{eqn_objective_function_coupled}, we normalized the unary and binary weights in each frame between $0$ and $1$. We use $\alpha=0.5$ in Eqn.\ref{eqn_detection_score}, and $\lambda_c=\lambda_s=\lambda_T=1$ for Eqn.\ref{eqn_objective_function_single},\ref{eqn_objective_function_coupled} and \ref{eqn_objective_function_pose_tree}. For $\sigma$ in Eqn.\ref{eqn_single_part_binary_weight} and $\theta$ in Eqn.\ref{eqn_coupled_part_unary_weight}, we use $10\%$ of the median height (normally 15-30 pixels) of N-Best poses \cite{Park2011} obtained from the step in Section \ref{sec_single_frame_hypotheses}. For each real body part (Section \ref{sec_single_frame_hypotheses}), we generate 20 hypotheses, and for each abstract body part we select the top 10 tracklets (Section \ref{sec_single_part_tracklets} and \ref{sec_coupled_part_tracklets}). In Eqn.\ref{eqn_intersection_penalty}, we use $c = 1/3$

\subsection{Computation Time}

We performed experiments on a desktop computer with Intel Core i7-3960X CPU at 3.3GHz and 16GB RAM. On average, to process one frame (typical frame size: $600 \times 800$, we resize the larger frames), the Matlab implementation took $0.5s$ to generate the body part hypotheses and weights, $0.5s$ to build the graph and compute the tracklets, and it took $0.1s$ to build the pose hypothesis graph (Section \ref{sec_optimal_pose_estimation}) and get the optimal solution. The limb alignment and refinement took less than $2s$ for 100 frames.

\subsection{Limitations and Failure Cases}

\begin{figure*}
\begin{center}
\includegraphics[width=7in]{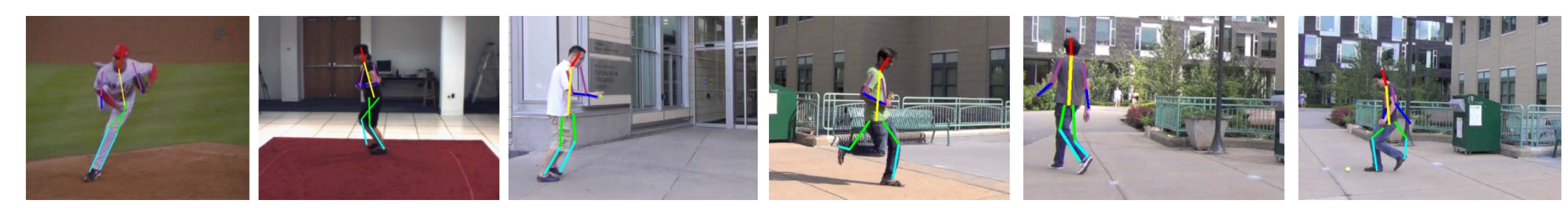}
   \caption{Failure Cases. Some of the failure cases are caused by occlusion (1-3), and some are caused by failed detection of joints (4-6). }
   \label{fig_result_failure_cases}
\end{center}
\end{figure*}

Although, the proposed method works quite well on the datasets, it still has some limitations and possible failure cases (please refer to Figure \ref{fig_result_failure_cases}). The main reason for the failure cases is that the proposed method relies on the pose hypothesis generation (i.e. using N-Best method \cite{Park2011} or DCNNGM \cite{Chen2014}) to generate multiple hypotheses for each body parts in each frame. And if the hypothesis generation method can not generate feasible body part hypotheses, the proposed method will fail. Figure \ref{fig_result_failure_cases} shows some failure cases (even though they may not be totally 'failed' but have larger errors) and they can be classified into three cases. 1) Occlusions (the left three images): if the body parts are occluded, image based pose estimation methods \cite{Park2011,Chen2014} can not predict the exact locations for both the body parts accurately. 2) One or more body parts are far away from the body center (the right three images): if the body parts are far away from the body center, they will be penalized by the spatial constraints from the graphical model in \cite{Park2011,Chen2014}. 3) The human pose in the frame is very different from the poses in the training set (image 1). In the training set of pose estimation model, most of the frames are captured from frontal views.

\section{Conclusions and future work}
\label{sec_conclusion}

We have proposed a tree-based optimization method for human pose estimation in videos. We have demonstrated that, by using the temporal information within the frames of a video, the performance of human pose estimation in videos can be significantly improved over the the image based pose estimation methods (even for deep learning methods). Our main contribution is  mostly focused on reformulating the problem to remove the simple cycles from the graph at the same time maintaining the useful connections at the greatest possible extent, in order to transform the original NP-hard problem into a simpler tree based optimization problem, for which the exact solution exists and can be solved efficiently. It is clear from the experiments that the proposed approach not only elegantly formulates the problem, but also dramatically improves the human pose estimation results in videos. The proposed formulation is general, it has a potential to be employed in solving some other problems in computer vision.


%

\ifCLASSOPTIONcaptionsoff
  \newpage
\fi



\bibliographystyle{IEEEtran}
%
\bibliography{R21}

%





\end{document}